\documentclass{article}

\PassOptionsToPackage{numbers, compress}{natbib}

\usepackage[preprint]{neurips_2021}

\usepackage[utf8]{inputenc} 
\usepackage[T1]{fontenc}    
\usepackage{hyperref}       
\usepackage{url}            
\usepackage{booktabs}       
\usepackage{amsfonts}       
\usepackage{nicefrac}       
\usepackage{microtype}      
\usepackage{xcolor}         

\usepackage{graphicx}
\usepackage{subfigure}
\usepackage{algorithm}
\usepackage{algorithmic}
\usepackage{amsmath,amssymb}
\usepackage{amsthm}
\usepackage{enumitem}
\usepackage{comment}
\usepackage{bm}
\usepackage{bbm}
\usepackage{makecell}
\usepackage{wrapfig}

\newcommand{\Set}[1]{{\mathcal{#1}}}

\newcommand{\E}[0]{\mathbb{E}}
\renewcommand{\vec}[1]{{\bm{#1}}}

\DeclareMathOperator*{\argmax}{arg\,max}

\makeatletter
\newcommand{\specificthanks}[1]{\@fnsymbol{#1}}
\makeatother

\newcommand\fullname{{\textbf{FAC}tored \textbf{M}ulti-\textbf{A}gent \textbf{C}entralised policy gradients}}
\newcommand\name{{FACMAC}}

\title{FACMAC: Factored Multi-Agent Centralised \\ Policy Gradients}

\author{%
Bei Peng\thanks{Equal contribution. Correspondence to: Bei Peng <\href{mailto:bei.peng@cs.ox.ac.uk}{bei.peng@cs.ox.ac.uk}>}
\textsuperscript{ \specificthanks{2}}
\And
Tabish Rashid$^*$\thanks{University of Oxford}
\And
Christian A. Schroeder de Witt$^*$$^\dagger$
\And
Pierre-Alexandre Kamienny\thanks{Facebook AI Research}
\And
Philip H. S. Torr$^\dagger$
\And
Wendelin B\"{o}hmer\thanks{Delft University of Technology}
\And
Shimon Whiteson$^\dagger$
}

\begin{document}

\maketitle

\begin{abstract}
We propose \fullname{} (\name{}), a new method for cooperative multi-agent reinforcement learning in both discrete and continuous action spaces.  Like MADDPG, a popular multi-agent actor-critic method, our approach uses deep deterministic policy gradients to learn policies.  However, \name{} learns a centralised but factored critic, which combines per-agent utilities into the joint action-value function via a non-linear monotonic function, as in QMIX, a popular multi-agent $Q$-learning algorithm. However, unlike QMIX, there are no inherent constraints on factoring the critic.
We thus also employ a nonmonotonic factorisation and empirically demonstrate that its increased representational capacity allows it to solve some tasks that cannot be solved with monolithic, or monotonically factored critics. In addition, \name{} uses a centralised policy gradient estimator that optimises over the entire joint action space, rather than optimising over each agent’s action space separately as in MADDPG.
This allows for more coordinated policy changes and fully reaps the benefits of a centralised critic. We evaluate \name{} on variants of the multi-agent particle environments, a novel multi-agent MuJoCo benchmark, and a challenging set of StarCraft II micromanagement tasks.
Empirical results demonstrate \name{}’s superior performance over MADDPG and other baselines on all three domains.
\end{abstract}

\section{Introduction}
Significant progress has been made in cooperative multi-agent reinforcement learning (MARL) under the paradigm of \textit{centralised training with decentralised execution} (CTDE) \cite{oliehoek_optimal_2008,kraemer_multi-agent_2016} in recent years, both in value-based \cite{sunehag2018value,rashid2018qmix,son2019qtran,rashid2020weighted,wang2020qplex,pan2021softmax} and actor-critic \cite{lowe2017multi,foerster_counterfactual_2018,iqbal2019actor,du_liir:_2019} approaches. Most popular multi-agent actor-critic methods such as COMA \cite{foerster_counterfactual_2018} and MADDPG \cite{lowe2017multi} learn a \textit{centralised critic} with decentralised actors. The critic is centralised to make use of all available information (i.e., it can condition on the global state and the joint action) to estimate the joint action-value function $Q_{tot}$, unlike a \textit{decentralised critic} that estimates the local action-value function $Q_a$ based only on individual observations and actions for each agent $a$.\footnote{COMA learns a single centralised critic for all cooperative agents due to parameter sharing. For each agent the critic has different inputs and can thus output different values for the same state and joint action. In MADDPG, each agents learns its own centralised critic, as it is designed for general multi-agent learning problems, including cooperative, competitive, and mixed settings.} 
Even though the joint action-value function these actor-critic methods can represent is not restricted, in practice they significantly underperform value-based methods like QMIX \cite{rashid2018qmix} on the challenging StarCraft Multi-Agent Challenge (SMAC) \cite{samvelyan19smac} benchmark \cite{rashid2020monotonic,rashid2020weighted}.

In this paper, we propose a novel approach called \textit{\fullname{}} (\name{}), which works for both discrete and continuous cooperative multi-agent tasks.
Like MADDPG, our approach uses deep deterministic policy gradients \cite{lillicrap_continuous_2015} to learn decentralised policies. 
However, \name{} learns a single \textit{centralised but factored critic}, which factors the joint action-value function $Q_{tot}$ into per-agent utilities $Q_a$ that are combined via a non-linear monotonic function, as in the popular $Q$-learning algorithm QMIX \citep{rashid2018qmix}. While the critic used in COMA and MADDPG is also centralised, it is monolithic rather than factored.\footnote{We use ``centralised and monolithic critic'' and ``monolithic critic'' interchangeably to refer to the centralised critic used in COMA and MADDPG, and ``centralised but factored critic'' and ``factored critic'' interchangeably to refer to the critic used in our approach.} Compared to learning a monolithic critic, our factored critic can potentially scale better to tasks with a larger number of agents and/or actions. In addition, in contrast to other value-based approaches such as QMIX, there are no inherent constraints on factoring the critic. This allows us to employ rich value factorisations, including \textit{nonmonotonic} ones, that value-based methods cannot directly use without forfeiting decentralisability or introducing other significant algorithmic changes. We thus also employ a nonmonotonic factorisation and empirically demonstrate that its increased representational capacity allows it to solve some tasks that cannot be solved with monolithic, or monotonically factored critics. 

In MADDPG, a separate policy gradient is derived for each agent individually, which optimises its policy assuming all other agents' actions are fixed. This could cause the agents to converge to sub-optimal policies in which no single agent wishes to change its action unilaterally. In \name{}, we use a new \textit{centralised} gradient estimator that optimises over the entire joint action space, rather than optimising over each agent's action space separately as in MADDPG. The agents' policies are thus trained as a single joint-action policy, which can enable learning of more coordinated behaviour, as well as the ability to escape  sub-optimal solutions. The centralised gradient estimator fully reaps the benefits of learning a centralised critic, by not implicitly marginalising over the actions of the other agents in the policy-gradient update. The gradient estimator used in MADDPG is also known to be vulnerable to relative overgeneralisation \cite{wei2016lenient}. To overcome this issue, in our centralised gradient estimator, we sample all actions from all agents' current policies when evaluating the joint action-value function. We empirically show that MADDPG can quickly get stuck in local optima in a simple continuous matrix game, whereas our centralised gradient estimator finds the optimal policy. While \citet{lyu2021contrasting} recently show that merely using a centralised critic (with per-agent gradients that optimise over each agent's actions separately) does not necessarily lead to better coordination, our centralised gradient estimator re-establishes the value of using centralised critics.

Most recent works on continuous MARL focus on evaluating their algorithms on the multi-agent particle environments \cite{lowe2017multi}, which feature a simple two-dimensional world with some basic simulated physics. To demonstrate \name{}'s scalability to more complex continuous domains and to stimulate more progress in continuous MARL, we introduce \textit{Multi-Agent MuJoCo} (MAMuJoCo), a new, comprehensive benchmark suite that allows the study of decentralised continuous control. 
Based on the popular single-agent MuJoCo benchmark \citep{brockman_openai_2016}, MAMuJoCo features a wide variety of novel robotic control tasks in which multiple agents within a single robot have to solve a task cooperatively. 

We evaluate \name{} on variants of the multi-agent particle environments \cite{lowe2017multi} and our novel MAMuJoCo benchmark, which both feature continuous action spaces, and the challenging SMAC benchmark \cite{samvelyan19smac}, which features discrete action spaces. Empirical results demonstrate \name{}'s superior performance over MADDPG and other baselines on all three domains. In particular, \name{} scales better when the number of agents (and/or actions) and the complexity of the task increases. Results on SMAC show that \name{} significantly outperforms stochastic DOP \cite{wang2021dop}, which recently claimed to be the first multi-agent actor-critic method to outperform state-of-the-art valued-based methods on SMAC, in all scenarios we tested. Moreover, our ablations and additional experiments demonstrate the advantages of both factoring the critic and using our centralised gradient estimator. We show that, compared to learning a monolithic critic, learning a factored critic can: 1) better take advantage of the centralised gradient estimator to optimise the agent policies when the number of agents and/or actions is large, and 2) leverage a nonmonotonic factorisation to solve tasks that cannot be solved with monolithic or monotonically factored critics. 
\section{Background} 
We consider a {\em fully cooperative multi-agent task} in which a team of agents interacts with the same environment to achieve some common goal. 
It can be modeled as a {\em decentralised partially observable Markov decision process} (Dec-POMDP) \cite{oliehoek_concise_2016} consisting of a tuple $G=\left\langle \mathcal{N},S,U,P,r,\Omega,O,\gamma\right\rangle$.
Here $ \mathcal{N} \equiv \{1,\dots,n\}$ denotes the finite set of agents and $s \in S$ describes the true state of the environment. 
At each time step, each agent $a \in \mathcal{N}$ selects a discrete or continuous action $u_a\in U$, forming a joint action $\mathbf{u}\in\mathbf{U}\equiv U^n$. 
This results in a transition to the next state $s'$ according to the state transition function $P(s'|s,\mathbf{u}):S\times\mathbf{U}\times S\rightarrow [0,1]$ and a team reward $r(s,\mathbf{u})$. $\gamma \in [0,1)$ is a discount factor.
Due to the \textit{partial observability}, each agent $a \in \mathcal{N}$ draws an individual partial observation $o_a \in \Omega$ from the observation kernel $O(s,a)$. 
Each agent learns a stochastic policy $\pi_a(u_a|\tau_a)$ or a deterministic policy $\mu_a(\tau_a)$, conditioned only on its local action-observation history $\tau_a\in T\equiv(\Omega \times U)^*$.
The joint stochastic policy $\boldsymbol{\pi}$ induces a joint {\em action-value function}: $Q^{\boldsymbol{\pi}}(s_t, \mathbf{u}_t)=\mathbb{E}_{s_{t+1:\infty}, \mathbf{u}_{t+1:\infty}} \left[R_t|s_t,\mathbf{u}_t\right]$, where $R_t=\sum^{\infty}_{i=0}\gamma^ir_{t+i}$ is the discounted return. Similarly, the joint deterministic policy $\boldsymbol{\mu}$ induces a joint action-value function denoted $Q^{\boldsymbol{\mu}}(s_t, \mathbf{u}_t)$.
We adopt the \textit{centralised training with decentralised execution} (CTDE) paradigm \cite{oliehoek_optimal_2008,kraemer_multi-agent_2016}, where policy training can exploit extra global information that might be available and has the freedom to share information between agents during training. However, during execution, each agent must act with only access to its own action-observation history. 

\paragraph{VDN and QMIX.} 
VDN \cite{sunehag2018value} and QMIX \cite{rashid2018qmix} are $Q$-learning algorithms for cooperative MARL tasks with discrete actions.
They both aim to efficiently learn a centralised but factored action-value function $Q_{tot}^{\boldsymbol{\pi}}$, using CTDE. 
To ensure consistency between the centralised and decentralised policies,
VDN and QMIX factor $Q_{tot}^{\boldsymbol{\pi}}$ assuming additivity and monotonicity, respectively.
More specifically, VDN factors $Q_{tot}^{\boldsymbol{\pi}}$ into a sum of the per-agent utilities: 
$Q_{tot}^{\boldsymbol{\pi}}(\boldsymbol{\tau}, \mathbf{u};\boldsymbol{\phi}) = \sum_{a=1}^{n}Q_a^{\pi_a}(\tau_a, u_a;\phi_a)$.
QMIX, however, represents $Q_{tot}^{\boldsymbol{\pi}}$ as a continuous monotonic mixing function of each agent's utilities: $Q_{tot}^{\boldsymbol{\pi}}(\boldsymbol{\tau}, \mathbf{u}, s;\boldsymbol{\phi}, \psi) = f_{\psi}\bigl(s,Q_1^{\pi_1}(\tau_1,u_1;\phi_1), \dots, Q_n^{\pi_n}(\tau_n,u_n;\phi_n) \bigr)$, where
$\frac{\partial f_{\psi}}{\partial Q_a} \geq 0, \forall a \in \mathcal{N}$.
This is sufficient to ensure that the global $\argmax$ performed on $Q_{tot}^{\boldsymbol{\pi}}$ yields the same result as a set of individual $\argmax$ performed on each $Q_{a}^{\pi_a}$.
Here $f_{\psi}$ is approximated by a monotonic mixing network, parameterised by $\psi$. 
Monotonicity can be guaranteed by non-negative mixing weights. These weights are generated by separate {\em hypernetworks} \citep{ha_hypernetworks_2016}, which condition on the full state $s$. 
QMIX is trained end-to-end to minimise the following loss:
\begin{equation}
\label{eq:qmix_loss}
	\mathcal{L}(\boldsymbol{\phi}, \psi) =  
	\E_{\Set D}\Big[ \big( y^{tot} 
	- Q_{tot}^{\boldsymbol{\pi}}(\boldsymbol{\tau},\mathbf{u}, s; \boldsymbol{\phi},\psi) \big)^2 \Big],  
\end{equation}
where the bootstrapping target $y^{tot}=r + \gamma \max_{\mathbf{u'}}Q_{tot}^{\boldsymbol{\pi}}(\boldsymbol{\tau'},\mathbf{u'}, s'; \boldsymbol{\phi^{-}}, \psi^{-})$. Here $r$ is the global reward, and $\boldsymbol{\phi^{-}}$ and $\psi^{-}$ are parameters of the target $Q$ and mixing network, respectively, as in DQN \cite{mnih2015human}. The expectation is estimated with a minibatch of transitions sampled from an experience replay buffer $\Set D$ \cite{lin_self-improving_1992}. 
During execution, each agent selects actions greedily with respect to its own $Q_{a}^{\pi_a}$. 

\paragraph{MADDPG.} MADDPG \cite{lowe2017multi} is an extension of DDPG \citep{lillicrap_continuous_2015} to multi-agent settings. It is an actor-critic, off-policy method that uses the paradigm of CTDE to learn deterministic policies in continuous action spaces. In MADDPG, a separate actor and critic is learned for each agent, such that arbitrary reward functions can be learned. It is therefore applicable to either cooperative, competitive, or mixed settings.
We assume each agent $a$ has a deterministic policy $\mu_a(\tau_a; \theta_a)$, parameterised by $\theta_a$ (abbreviated as $\mu_a$), and let
$\boldsymbol{\mu}=\{\mu_a(\tau_a;\theta_a)\}_{a=1}^{n}$ be the set of all agent policies.
In MADDPG, a {\em centralised and monolithic critic} that estimates the joint action-value function $Q_{a}^{\boldsymbol{\mu}}(s,u_1,\dots,u_n;\phi_a)$
is learned for each agent $a$ separately.
The critic is said to be \textit{centralised} as it utilises information only available to it during the \textit{centralised} training phase, the global state $s$\footnote{If the global state $s$ is not available, the centralised and monolithic critic can condition on the joint observations or action-observation histories.} and the actions of all agents, $u_1,\dots,u_n$, to estimate the joint action-value function $Q_{a}^{\boldsymbol{\mu}}$, which is parameterised by $\phi_a$. 
This joint action-value function is trained by minimising the following loss:
\begin{equation}	 
	\mathcal{L}(\phi_a) = 
		\E_{\Set D}\Big[
		\Big(y^{a} - Q_a^{\boldsymbol{\mu}}(s, u_1,\dots,u_n; \phi_a)
		\Big)^{2} \Big], 
\end{equation} 
where $y^{a}=r_a + \gamma Q_a^{\boldsymbol{\mu}}(s', u_{1}',\dots,u_{n}'|_{u_{a}'=\mu_{a}(\tau_a;\theta_{a}^{-})};\phi_{a}^{-})$. Here $r_a$ is the reward received by each agent $a$, $u_{1}',\dots,u_{n}'$ is the set of target policies with delayed parameters $\theta_{a}^{-}$, and $\phi_{a}^{-}$ are the parameters of the target critic. The replay buffer $\Set D$ contains the transition tuples $(s,s',u_1,\dots,u_n,r_1,\dots,r_n)$. 

The following policy gradient can be calculated individually to update the policy of each agent $a$:
\begin{equation} 
\label{eq:dpg}
	\nabla_{\theta_a} J(\mu_a) =
    \E_{\Set D}\Big[ \nabla_{\theta_a}\mu_a(\tau_a) 
		\nabla_{u_{a}}Q_a^{\boldsymbol{\mu}}(s,u_1,\dots,u_n)\big|_{u_{a}=\mu_{a}(\tau_a)} \Big],
\end{equation}
where the current agent $a$’s action $u_a$ is sampled from its current policy $\mu_a$ when evaluating the joint action-value function $Q_a^{\boldsymbol{\mu}}$, while all other agents’ actions are sampled from the replay buffer $\Set D$. 
\section{\name{}}
In this section, we propose a new approach called \textit{\fullname{}} (\name{}) that uses a centralised but factored critic and a centralised gradient estimator to learn continuous cooperative tasks. We start by describing the idea of learning a centralised but factored critic. We then discuss our new centralised gradient estimator and demonstrate its benefit in a simple continuous matrix game. Finally, we discuss how we adapt our method to discrete cooperative tasks.

\subsection{Learning a Centralised but Factored Critic}
\label{sec: factored_critic}
Learning a centralised and monolithic critic conditioning on the global state and the joint action can be difficult and/or impractical when the number of agents and/or actions is large \citep{iqbal2019actor}. 
We thus employ value function factorisation in the multi-agent actor-critic framework to enable scalable learning of a centralised critic in Dec-POMDPs. 
Another key advantage of adopting value factorisation in an actor-critic framework is that, compared to value-based methods, it allows for a more flexible factorisation as the critic's design is not constrained.
One can employ any type of factorisation, including nonmonotonic factorisations that value-based methods cannot directly use without forfeiting decentralisability or introducing other significant algorithmic changes.

Specifically, in \name{}, all agents share a centralised critic $Q_{tot}^{\boldsymbol{\mu}}$ that is factored as:
\begin{align}
    Q_{tot}^{\boldsymbol{\mu}}(\boldsymbol{\tau}, \mathbf{u}, s;\boldsymbol{\phi},\psi) = g_{\psi}\bigl(s,\{ Q_a^{\mu_a}(\tau_a,u_a;\phi_a) \}_{a=1}^{n} \bigr),
\end{align}
where $\boldsymbol{\phi}$ and $\phi_a$ are parameters of the joint action-value function $Q_{tot}^{\boldsymbol{\mu}}$ and agent-wise utilities $Q_a^{\mu_a}$, respectively.
In our canonical implementation which we refer to as \name{}, $g_\psi$ is a non-linear monotonic function parametrised as a mixing network with parameters $\psi$, as in QMIX \cite{rashid2018qmix}. 
To evaluate the policy, the centralised but factored critic is trained by minimising the following loss:
\begin{align}	 
	\mathcal{L}(\boldsymbol{\phi},\psi) = 
		\E_{\Set D}\Big[
		\Big(y^{tot} - Q_{tot}^{\boldsymbol{\mu}}(\boldsymbol{\tau}, \mathbf{u},s; \boldsymbol{\phi}, \psi)
		\Big)^{\!2} 
	\,\Big], 
\end{align} 
where $y^{tot}=r + \gamma Q_{tot}^{\boldsymbol{\mu}}(\boldsymbol{\tau'}, \boldsymbol{\mu}(\boldsymbol{\tau'};\boldsymbol{\theta^{-}}), s'; \boldsymbol{\phi^{-}}, \psi^{-})$.
Here $\Set D$ is the replay buffer, and $\boldsymbol{\theta^{-}}$, $\boldsymbol{\phi^{-}}$, and $\psi^{-}$ are parameters of the target actors, critic, and mixing network, respectively.

Leveraging the flexibility of our approach, namely the lack of restrictions on the form of the critic, we also explore a new nonmonotonic factorisation with full representational capacity. The joint action-value function $Q_{tot}^{\boldsymbol{\mu}}$ is represented as a non-linear non-monotonic mixing function of per-agent utilities $Q_a^{\mu_a}$. This nonmonotonic mixing function is parameterised as a mixing network, with a similar architecture to $g_\psi$ in \name{}, but without the constraint of monotonicity enforced by using non-negative weights. We refer to this method as \name{}-nonmonotonic. Additionally, to better understand the advantages of factoring a centralised critic, we also explore two additional simpler factorisation schemes.
These include factoring the centralised critic $Q_{tot}^{\boldsymbol{\mu}}$ into a sum of per-agent utilities $Q_a^{\mu_a}$ as in VDN (\name{}-vdn), and as a sum of $Q_a^{\mu_a}$ and a state-dependent bias (\name{}-vdn-s).
Our value factorisation technique is general and can be readily applied to any multi-agent actor-critic algorithms that learn centralised and monolithic critics \cite{lowe2017multi,foerster_counterfactual_2018,du_liir:_2019}.

\subsection{Centralised Policy Gradients}
\label{sec: cpg}
To update the decentralised policy of each agent, a naive adaptation of the deterministic policy gradient used by MADDPG (shown in \eqref{eq:dpg}) is 
\begin{align}
    \nabla_{\theta_a} J(\mu_a) =
	\E_{\Set D}\Big[\nabla_{\theta_a} \mu_a(\tau_a) 
		\nabla_{u_{a}}Q_{tot}^{\boldsymbol{\mu}}(\boldsymbol{\tau}, u_1,\dots,u_n,s)\big|_{u_{a}=\mu_{a}(\tau_a)}  \Big].
	\label{eq:dpg2}
\end{align}
Compared to the policy gradient used in MADDPG, the updates of all agents' individual deterministic policies now depend on the single shared factored critic $Q_{tot}^{\boldsymbol{\mu}}$, 
as opposed to learning and utilising a monolithic critic $Q_a^{\boldsymbol{\mu}}$ for each agent.
However, there are two main problems in both policy gradients. 
First, each agent optimises its own policy assuming all other agents' actions are fixed, which could cause the agents to converge to sub-optimal policies in which no single agent wishes to change its action unilaterally. 
Second, both policy gradients make the corresponding methods vulnerable to relative overgeneralisation \cite{wei2016lenient} as, when agent $a$ ascends the policy gradient based on $Q_a^{\boldsymbol{\mu}}$ or $Q_{tot}^{\boldsymbol{\mu}}$, only its own action $u_a$ is sampled from its current policy $\mu_a$, while all other agents' actions are sampled from the replay buffer $\Set D$. 
The other agents' actions thus might be drastically different from the actions their current policies would choose.
This could cause the agents to converge to sub-optimal actions that appear to be a better choice when considering the effect of potentially arbitrary actions from the other collaborating agents.

In \name{}, we use a new \textit{centralised} gradient estimator that optimises over the entire joint action space, rather than optimising over each agent's actions separately as in both \eqref{eq:dpg} and \eqref{eq:dpg2}, to achieve better coordination among agents. In addition, to overcome relative overgeneralisation, when calculating the policy gradient we sample all actions from all agents' current policies when evaluating $Q_{tot}^{\boldsymbol{\mu}}$. 
Our centralised policy gradient can thus be estimated as
\begin{align}
    \nabla_{\theta}J(\boldsymbol{\mu}) =
	\E_{\Set D}\Big[\nabla_{\theta} \boldsymbol{\mu}		\nabla_{\boldsymbol{\mu}}Q_{tot}^{\boldsymbol{\mu}}(\boldsymbol{\tau}, \mu_1(\tau_1), \dots, \mu_n(\tau_n), s) \Big],
\end{align}
where $\boldsymbol{\mu}=\{\mu_1(\tau_1; \theta_1), \dots, \mu_n(\tau_n;\theta_n) \}$ is the set of all agents' current policies and all agents share the same actor network parameterised by $\theta$. However, it is not a requirement of our method for all agents to share parameters in this manner.

If the critic factorisation is linear, as in \name{}-vdn, then the centralised gradient is equivalent to the per-agent gradients that optimise over each agent's actions separately. 
This is explored in more detail by DOP \citep{wang2021dop}, which restricts the factored critic to be linear to exploit this equivalence.
A major benefit of our method then, is that it does not place any such restrictions on the critic. As remarked by \citet{lyu2021contrasting}, merely using a centralised critic with per-agent gradients does not necessarily lead to better coordination between agents due to the two problems outlined above. Our centralised gradient estimator, which now optimises over the entire joint action space, is required in order to fully take advantage of a centralised critic. 

\begin{figure*}[t!]
    \centering
    \includegraphics[width=0.9\columnwidth]{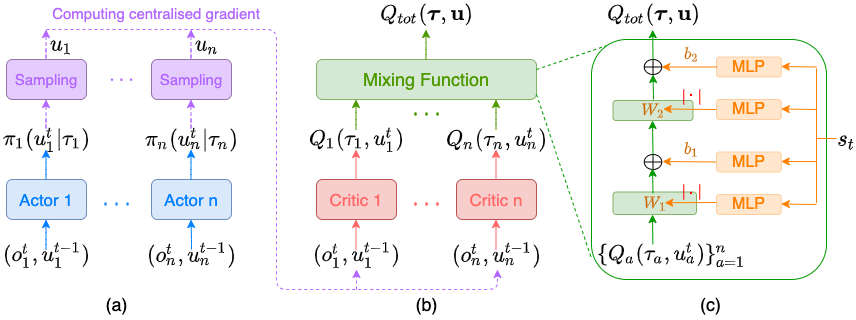}
    \caption{The overall \name{} architecture. (a) The decentralised policy networks. (b) The centralised but factored critic. (c) The non-linear monotonic mixing function.}
    \label{fig:factored_critic_arch}
\end{figure*}

Figure \ref{fig:factored_critic_arch} illustrates the overall \name{} architecture.
For each agent $a$, there is one policy network that decides which individual action (discrete or continuous) to take.
There is also one critic network for each agent $a$ that estimates the individual agent utilities $Q_a$, which are then combined into the joint action-value function $Q_{tot}$ via a non-linear monotonic mixing function as in QMIX. $Q_{tot}$ is then used by our centralised gradient estimator to help the actor update its policy parameters. 

\begin{figure}
    \centering
    \includegraphics[width=0.3\columnwidth]{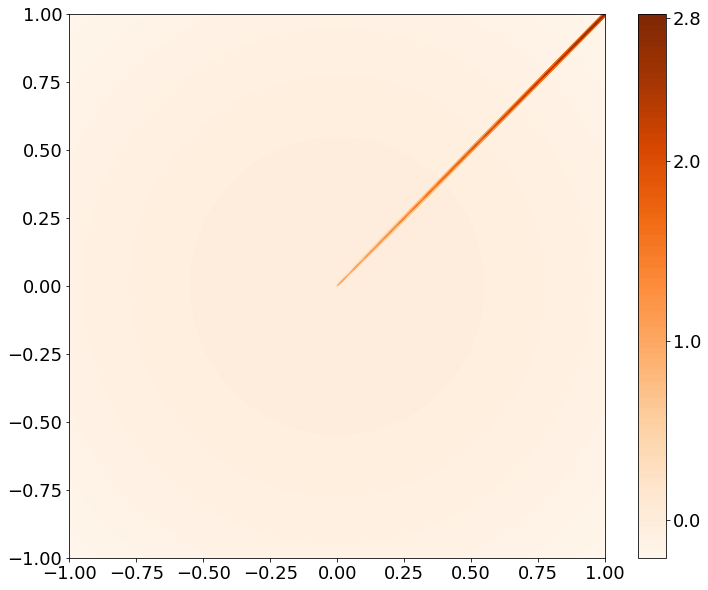}
    \includegraphics[width=0.35\columnwidth]{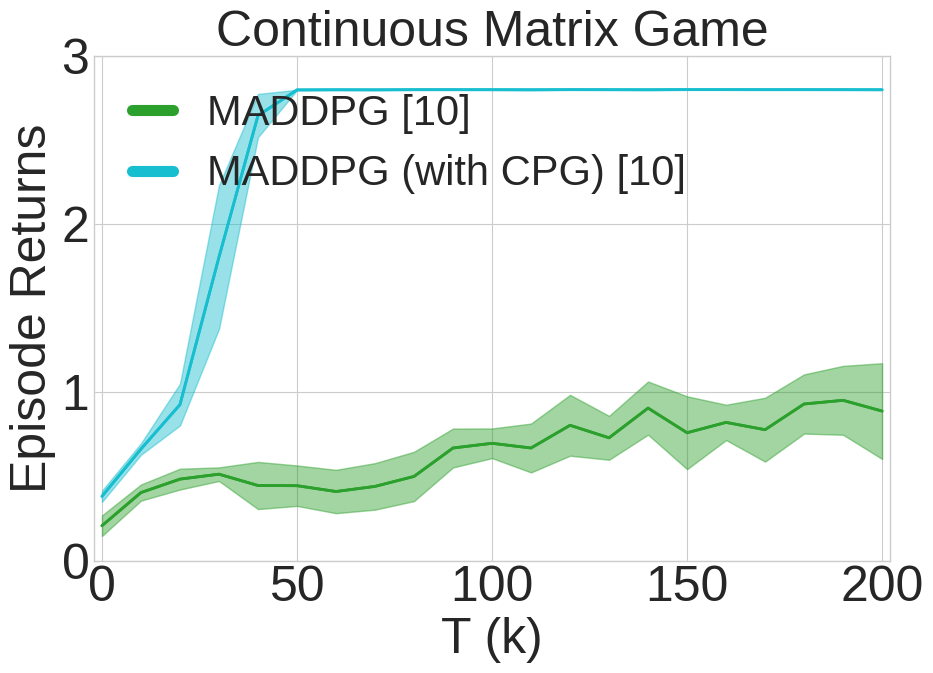}
    \caption{\textbf{Left:} The Continuous Matrix Game. \textbf{Right:} Mean test return on Continuous Matrix Game.}
    \label{fig:cts_matrix_game_result}
\end{figure}

\begin{figure}
    \centering
    \includegraphics[width=0.67\columnwidth]{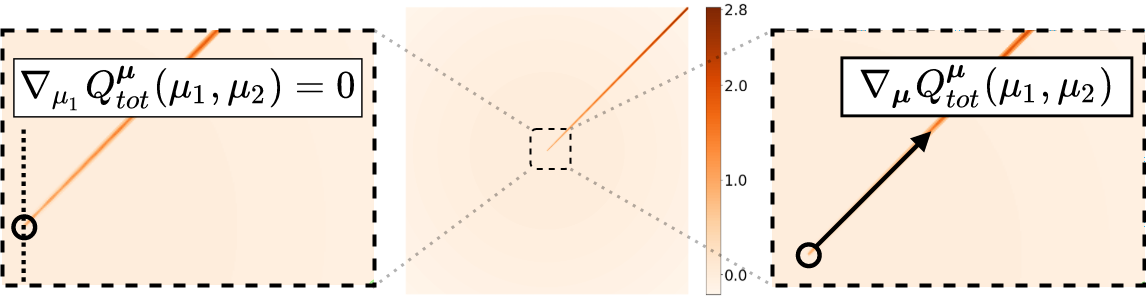}
    \caption{\textbf{Left:} Per-agent policy gradient at the origin. For agent 1 (similarly for agent 2) it is 0 since the gradient term assumes the other agent's action to be fixed and thus it only considers the relative improvements along the dotted line. \textbf{Right:} Our Centralised Policy Gradient correctly determines the gradient for improving the joint action.}
    \label{fig:cts_matrix_game_gradients}
\end{figure}

To show the benefits of our new centralised gradient estimator, we compare MADDPG with the centralised policy gradient (CPG) against the original MADDPG on a simple continuous cooperative matrix game.
Figure \ref{fig:cts_matrix_game_result} (left) illustrates the continuous matrix game with two agents. There is a narrow path (shown in red) starting from the origin $(0,0)$ to $(1,1)$, in which the reward gradually increases. 
Everywhere else there is a small punishment moving away from the origin, increasing in magnitude further from the origin.
Experimental results are shown in Figure \ref{fig:cts_matrix_game_result} (right). MADDPG quickly gets stuck in the local optimum within $200k$ timesteps, while MADDPG (with CPG)  robustly converges to the optimal policy. 
Figure \ref{fig:cts_matrix_game_gradients} visualises the differences between the per-agent and centralised policy gradients, demonstrating that the centralised policy gradient is necessary to take advantage of the centralised critic.
In Section \ref{sec: exps}, we further demonstrate the benefits of this centralised gradient estimator in more complex tasks. 

\subsection{Discrete Policy Learning}
As \name{} requires differentiable policies and the sampling process of discrete actions from a categorical distribution is not differentiable, we use the Gumbel-Softmax estimator \cite{jang2016categorical} to enable efficient learning of \name{} on cooperative tasks with discrete actions. The Gumbel-Softmax estimator is a continuous distribution that approximates discrete samples from a categorical distribution to produce differentiable samples. It is a differentiable relaxation of the Gumbel-Max trick, which reparameterises the stochastic policies as a deterministic function of the policy parameters and some independent noise sampled from a standard Gumbel distribution. 

Moreover, we use the Straight-Through Gumbel-Softmax Estimator \cite{jang2016categorical} to ensure the action dynamics during training and evaluation are the same. Specifically, during training, we sample discrete actions $u_a$ from the original categorical distribution in the forward pass, but use the continuous Gumbel-Softmax sample $x_a$ in the backward pass to approximate the gradients: $\nabla_{\theta_a}u_a \approx \nabla_{\theta_a}x_a$. 
We can then update the agent's policy using our centralised policy gradient: $\nabla_{\theta} J(\theta) \approx 
	\E_{\Set D}\Big[\nabla_{\theta} \boldsymbol{x} 
			\nabla_{\boldsymbol{x}}  Q_{tot}^{\boldsymbol{x}}(\boldsymbol{\tau}, x_1, \dots, x_n, s)
			\Big],$ 
where $\boldsymbol{x}=\{x_1, \dots, x_n\}$ is the set of continuous sample that approximates the discrete agent actions. 
\section{Multi-Agent MuJoCo}
\label{sec:mamujoco}
The evaluation of continuous MARL algorithms has recently been largely limited to the simple multi-agent particle environments \cite{lowe2017multi}. We believe the lack of diverse continuous benchmarks is one factor limiting progress in continuous MARL. To demonstrate \name{}'s scalability to more complex continuous domains and to stimulate more progress in continuous MARL, we develop \textit{Multi-Agent MuJoCo} (MAMuJoCo), a novel benchmark for continuous cooperative multi-agent robotic control. Starting from the popular fully observable single-agent robotic MuJoCo \citep{todorov_mujoco:_2012} control suite included with OpenAI Gym \citep{brockman_openai_2016}, we create a wide variety of novel scenarios in which multiple agents within a single robot have to solve a task cooperatively.

\begin{figure*}[t!]
	\begin{center}
	\includegraphics[width=0.98\textwidth]{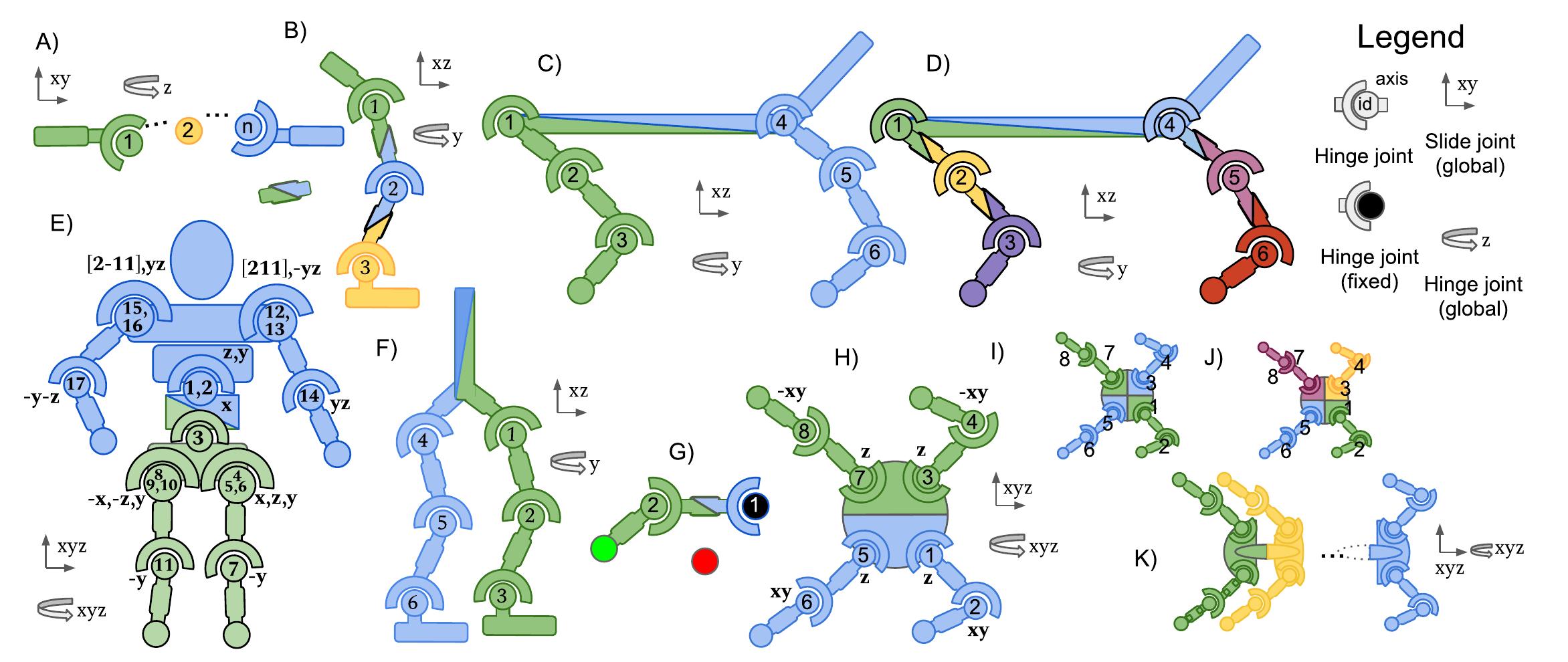}
	\end{center}
	\caption{\textbf{Agent partitionings for MAMuJoCo environments:} A) Manyagent Swimmer, B) 3-Agent Hopper [3x1], C) 2-Agent HalfCheetah  [2x3], D) 6-Agent HalfCheetah [6x1], E) 2-Agent Humanoid and 2-Agent HumanoidStandup (each [1x9,1x8]), F) 2-Agent Walker [2X3], G) 2-Agent Reacher [2x1], H) 2-Agent Ant [2x4], I) 2-Agent Ant Diag [2x4], J) 4-Agent Ant [4x2], and K) Manyagent Ant. Colours indicate agent partitionings. Each joint corresponds to a single controllable motor. Split partitions indicate shared body segments. Square brackets indicate [(number of agents) x (joints per agent)]. Joint IDs are in order of definition in the corresponding OpenAI Gym XML asset files \citep{brockman_openai_2016}. Global joints indicate degrees of freedom of the center of mass of the composite robotic agent.}
	\label{fig:multiagent_mujoco}
\end{figure*} 

Single-robot multi-agent tasks in MAMuJoCo arise by first representing a given single robotic agent as a \textit{body graph}, where vertices (joints) are connected by adjacent edges (body segments).
We then partition the body graph into disjoint sub-graphs, one for each agent, each of which contains one or more joints that can be controlled.
Figure \ref{fig:multiagent_mujoco} shows agent partitionings for MAMuJoCo environments.

Multiple agents are introduced within a single robot as partial observability arises through latency, bandwidth, and noisy sensors in a single robot. 
Even if communication is free and instant when it works, we want policies that keep working even when communication channels within the robot malfunction. 
Without access to the exact full state, local decision rules become more important and introducing autonomous agents at individual decision points (e.g., each physical component of the robot) is reasonable and beneficial. 
This also makes it more robust to single-point failures (e.g., broken sensors) and more adaptive and flexible as new independent decision points (thus agents) may be added easily. This design also offers important benefits. It facilitates comparisons to existing literature on both the fully observable single-agent domain \citep{openai_openaibaselines_2020}, as well as settings with low-bandwidth communication \citep{wang_nervenet_2018}.
More importantly, it allows for the study of novel MARL algorithms for decentralised coordination in isolation (scenarios with multiple robots may add confounding factors such as spatial exploration), which is currently a gap in the research literature.  

MAMuJoCo also includes scenarios with a larger and more flexible number of agents, which takes inspiration from modular robotics \citep{yim2002modular, kurokawa2008distributed}. Compared to traditional robots, modular robots are more versatile, configurable, and scalable as it is easier to replace or add modules to change the degrees of freedom. We therefore develop two scenarios named ManyAgent Swimmer and ManyAgent Ant, in which one can configure an arbitrarily large number of agents (within the memory limits), each controlling a consecutive segment of arbitrary length. This design is similar to many practical modular snake robots \citep{wright2012design, nakagaki_chainform_2016}, which mimic snake-like motion for diverse tasks such as navigating rough terrains and urban search and rescue. See Appendix \ref{sec:mamujoco_sup} for more details about MAMuJoCo.
\section{Experimental Results}
\label{sec: exps}
In this section we present our experimental results on our cooperative variants of the continuous \textit{simple tag} environment introduced by \citet{lowe2017multi} (we refer to this environment as Continuous Predator-Prey), our novel continuous benchmark MAMuJoCo, and the challenging SMAC\footnote{We utilise SC2.4.10., which is used by the latest PyMARL framework. The original results reported in \citet{samvelyan19smac} and \citet{rashid2020monotonic} use SC2.4.6. Performance is \textbf{not} always comparable across versions.} \cite{samvelyan19smac} benchmark with discrete action spaces. 
In discrete cooperative tasks, we compare with state-of-the-art multi-agent actor-critic algorithms MADDPG \cite{lowe2017multi}, COMA \cite{foerster_counterfactual_2018}, CentralV \cite{foerster_counterfactual_2018}, DOP \cite{wang2021dop}, and value-based methods QMIX \cite{rashid2018qmix} and QPLEX \cite{wang2020qplex}.
In continuous cooperative tasks, we compare with MADDPG \cite{lowe2017multi} and independent DDPG (IDDPG), as well as COVDN and COMIX, two novel baselines described below. We also explore different forms of critic factorisation. More details about the environments, experimental setup, and training details are included in Appendix \ref{sec:env_details} and \ref{sec:exp_details}.

\paragraph{COVDN and COMIX}
We find that not many multi-agent value-based methods work off the shelf with continuous actions. To compare \name{} against value-based approaches in continuous cooperative tasks, we use existing continuous $Q$-learning approaches in single-agent settings to extend VDN and QMIX to continuous action spaces. Specifically, we introduce COVDN and COMIX, which use VDN-style and QMIX-style factorisation respectively and both perform approximate greedy action selection using the cross-entropy method (CEM) \citep{de_boer_tutorial_2005}. CEM is a sampling-based derivative-free heuristic search method that has been successfully used to find approximate maxima of nonconvex $Q$-networks in single-agent robotic control tasks \citep{kalashnikov_qt-opt:_2018}. The centralised but factored $Q_{tot}$ allows us to use CEM to sample actions for each agent independently and to use the per-agent utility $Q_a$ to guide the selection of maximal actions.

In both COVDN and COMIX, CEM is used by each agent $a$ to find an action that approximately optimises its local utility function $Q_a$. Specifically, CEM iteratively draws a batch of $N$ random samples from a candidate distribution $\mathcal{D}_k$, e.g., a Gaussian, at each iteration $k$. The best $M < N$ samples (with the highest utility values) are then used to fit a new Gaussian distribution $\mathcal{D}_{k+1}$, and this process repeats $K$ times. We use a CEM hyperparameter configuration similar to Qt-Opt \citep{kalashnikov_qt-opt:_2018}, where $N=64$, $M=6$, and $K=2$.\footnote{We empirically find $2$ iterations to suffice.} Gaussian distributions are initialised with mean $\mu=0$ and standard deviation $\sigma=1$. Algorithm \ref{alg:cem} and \ref{alg:comix} in Appendix \ref{sec:comix} outline the full process for CEM and COMIX, respectively. We do not consider COVDN and COMIX significant algorithmic contributions but instead merely baseline algorithms.

\begin{figure*}[t!]
	\centering
	\subfigure[{3 agents and 1 prey}]{
		\label{fig:pp_3a}
		\includegraphics[width=0.31\textwidth]{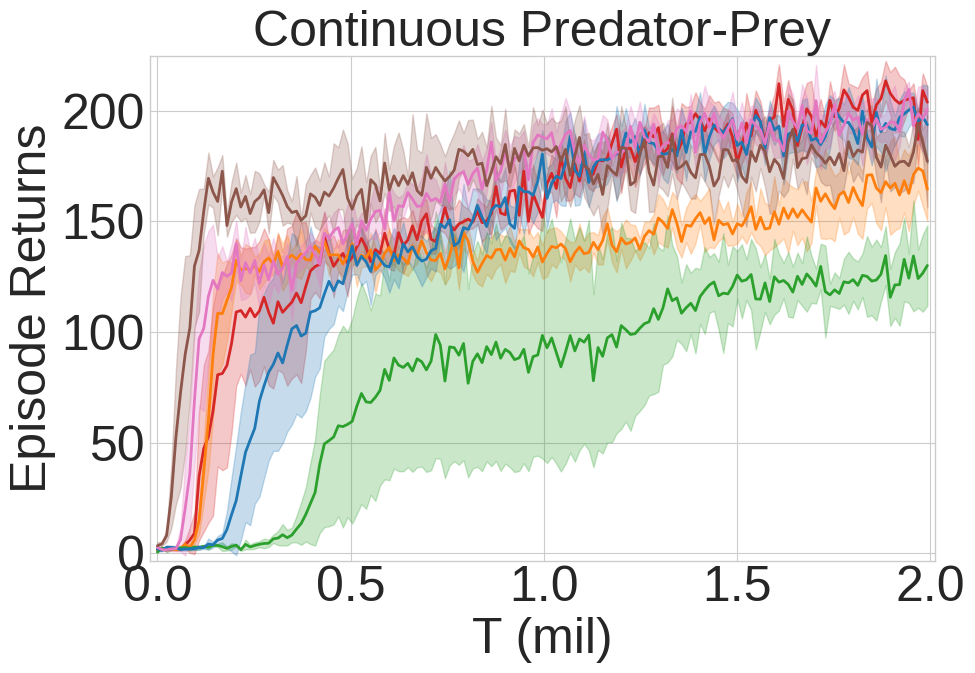}
	}
	\subfigure[{6 agents and 2 preys}]{
		\label{fig:pp_6a}
		\includegraphics[width=0.31\textwidth]{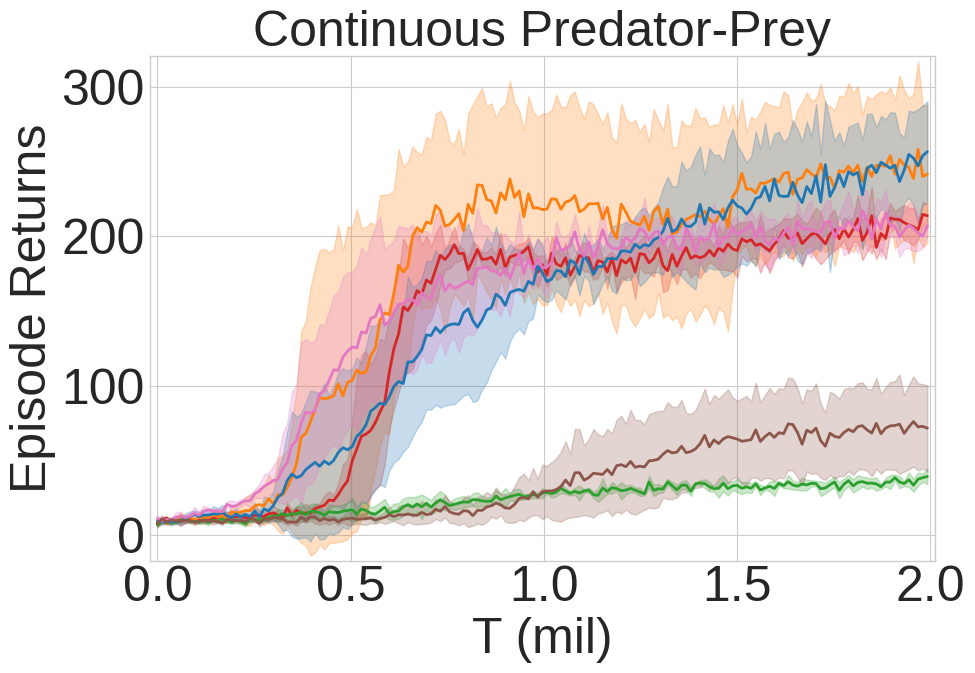}
	}
	\subfigure[{9 agents and 3 preys}]{
		\label{fig:pp_9a}
		\includegraphics[width=0.31\textwidth]{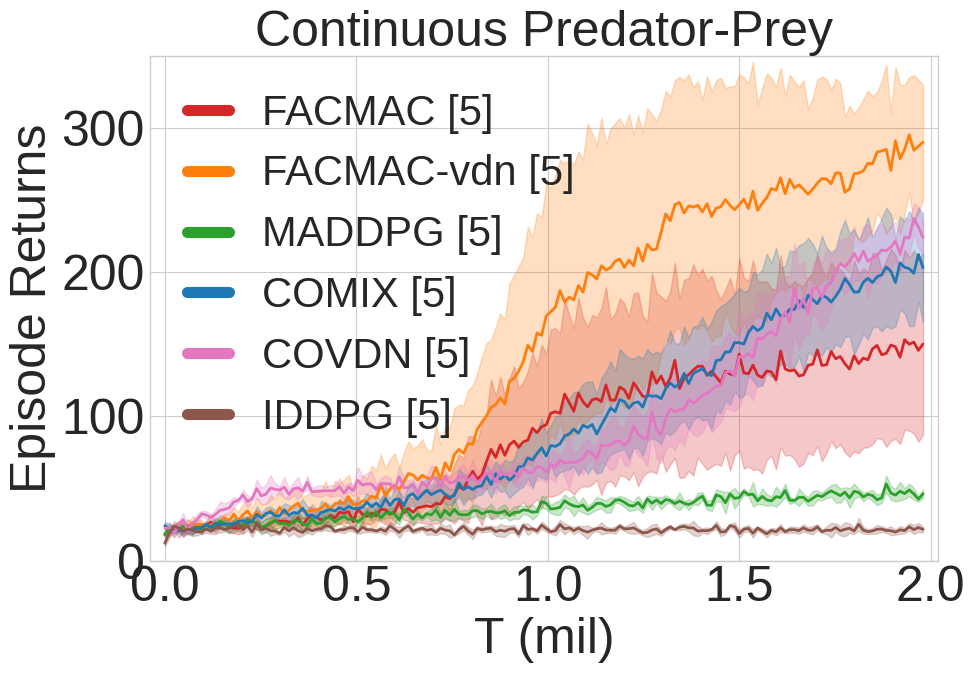}
	}
	\caption{Mean episode return on Continuous Predator-Prey with different number of agents and preys. The mean across $5$ seeds is plotted and the $95\%$ confidence interval is shown shaded.}
	\label{fig: pp_results}
\end{figure*}

\begin{figure*}[t!]
	\centering
    \includegraphics[width=0.32\textwidth]{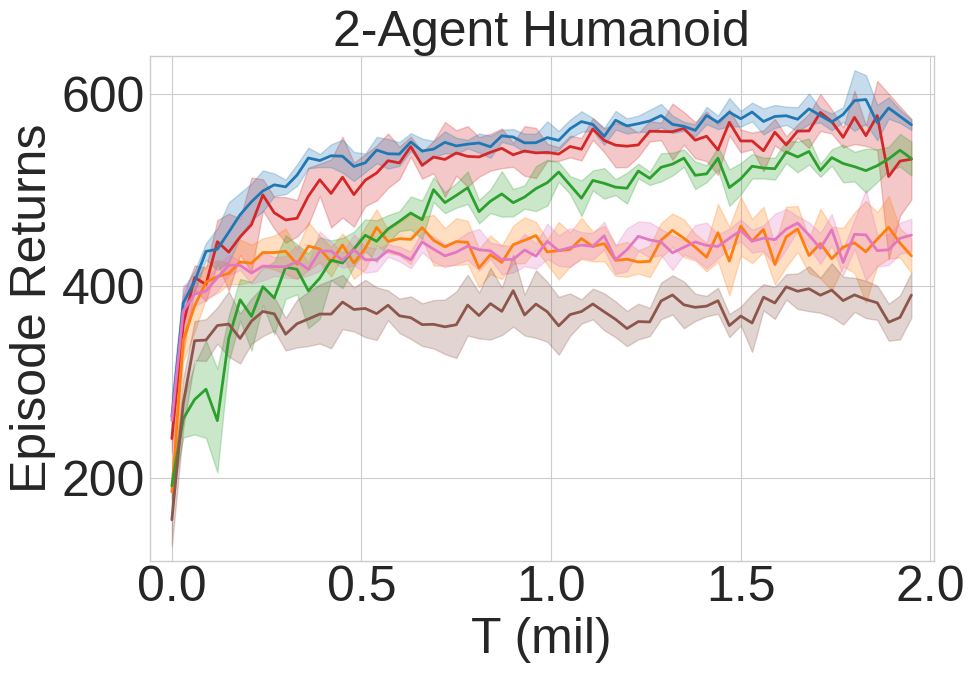}
	\includegraphics[width=0.32\textwidth]{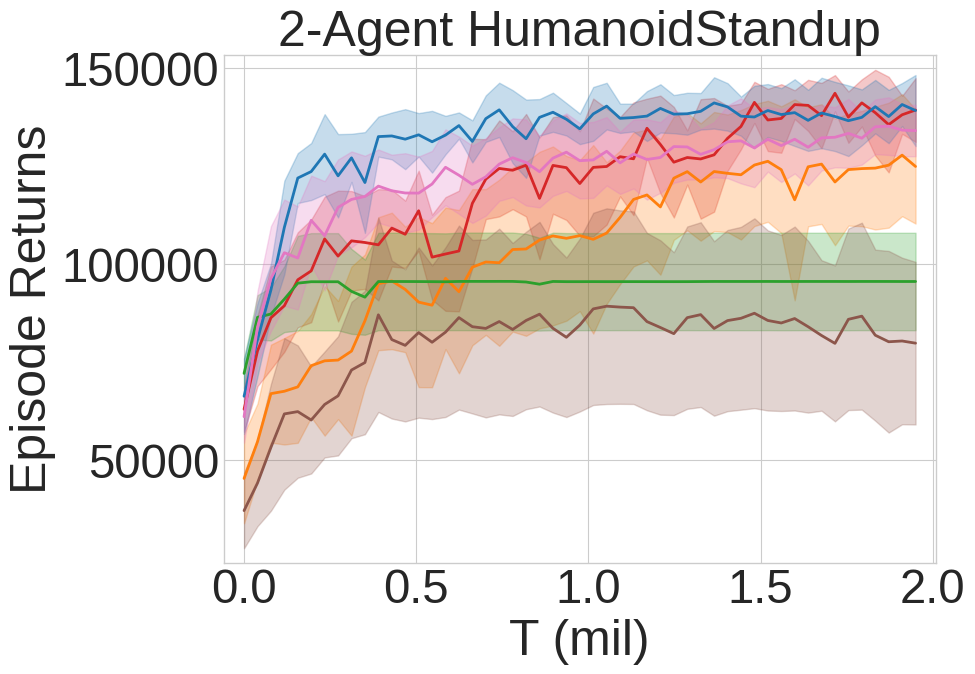}
	\includegraphics[width=0.32\textwidth]{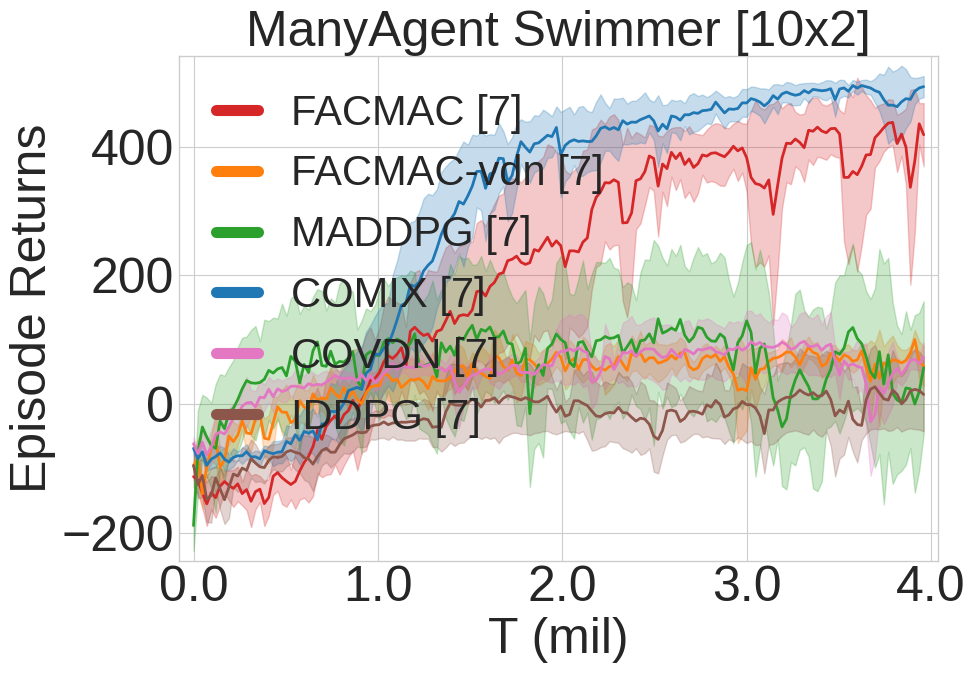}
	\caption{Mean episode return on different MAMuJoCo tasks. In ManyAgent Swimmer, we configure the number of agents to be $10$, each controlling a consecutive segment of length $2$. The mean across $7$ seeds is plotted and the $95\%$ confidence interval is shown shaded.}
	\label{fig: mamujoco_results}
\end{figure*}

\begin{figure*}[t!]
	\centering
	\subfigure[{Easy}]{
		\label{fig:smac_2s3z}
		\includegraphics[width=0.3\textwidth]{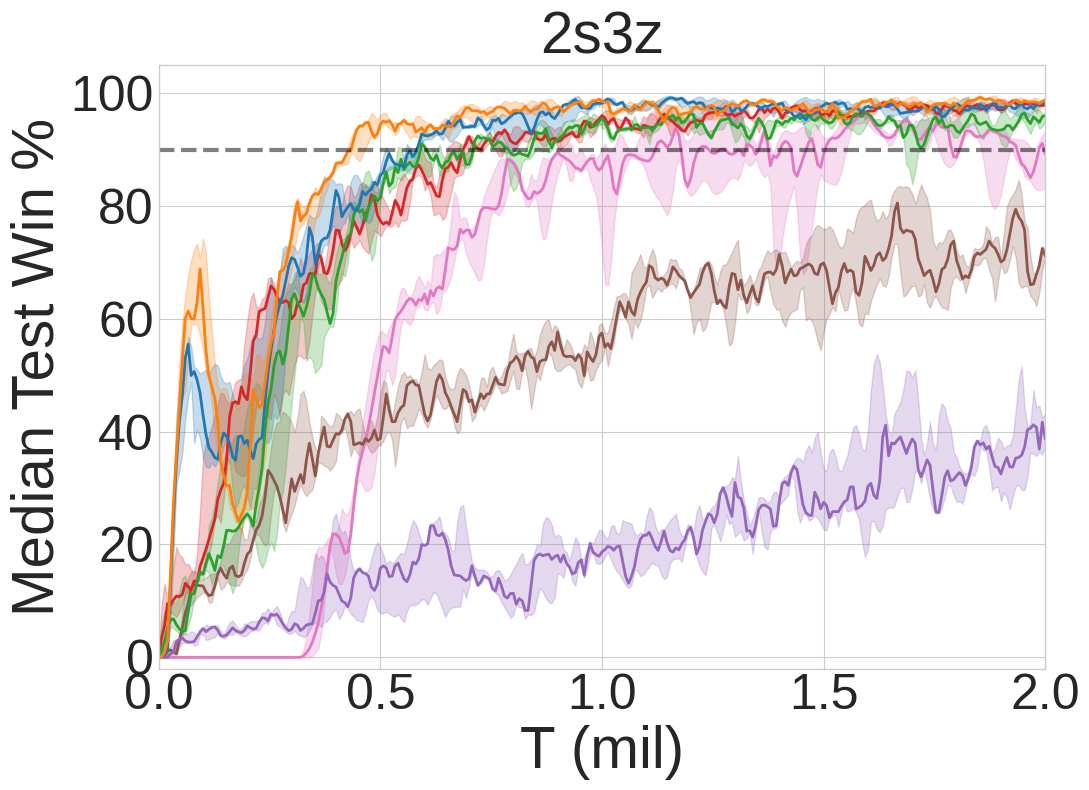}
	}
	\subfigure[{Easy}]{
		\label{fig:smac_mmm}
		\includegraphics[width=0.3\textwidth]{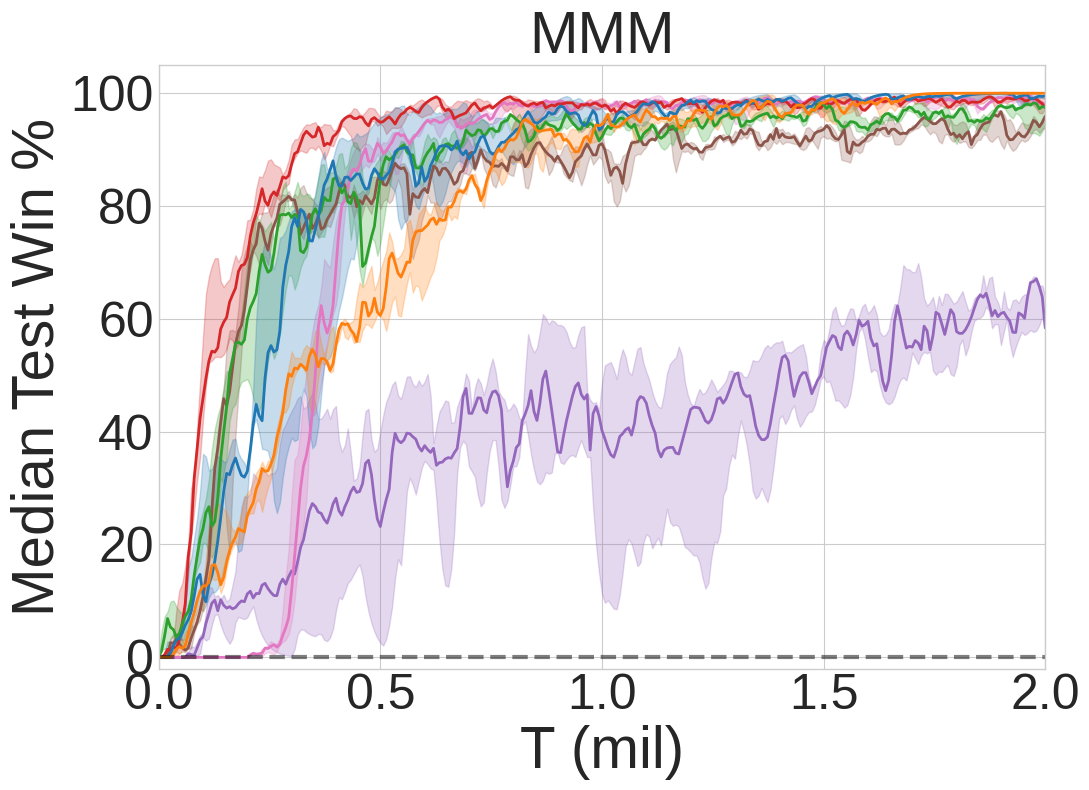}
	}
	\subfigure[{Hard}]{
		\label{fig:smac_2c_vs_64zg}
		\includegraphics[width=0.3\textwidth]{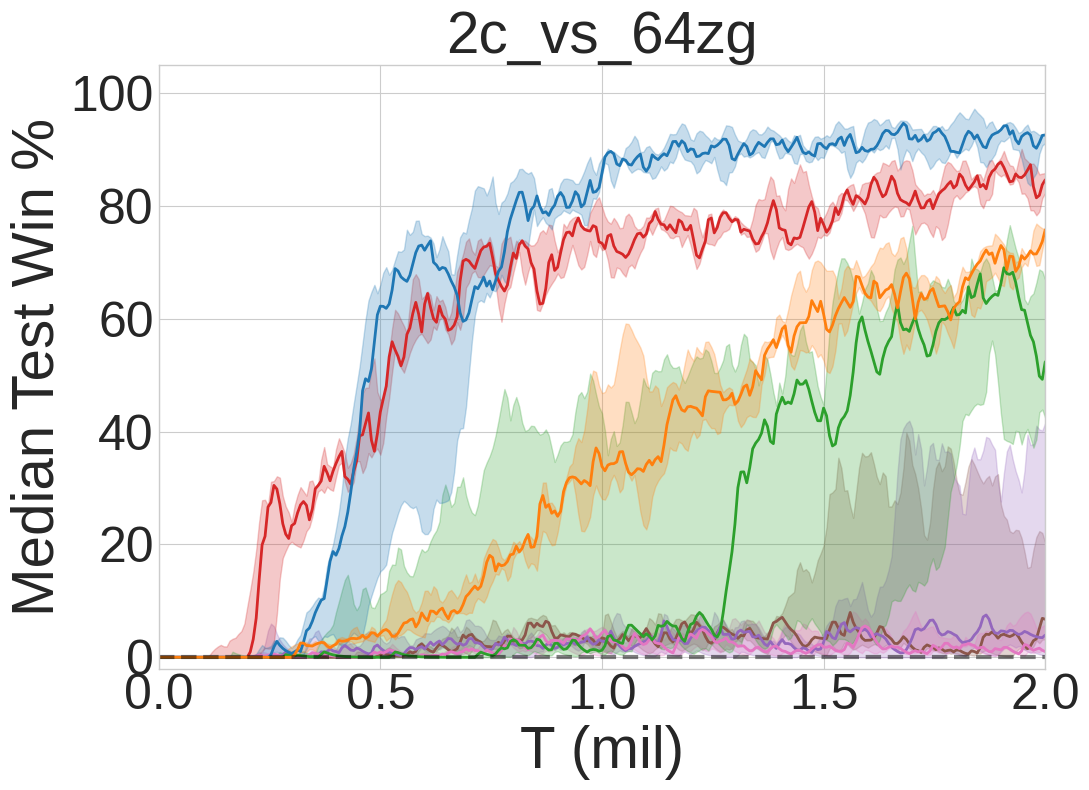}
	}
	\subfigure[{Hard}]{
		\label{fig:smac_bane_vs_bane}
		\includegraphics[width=0.3\textwidth]{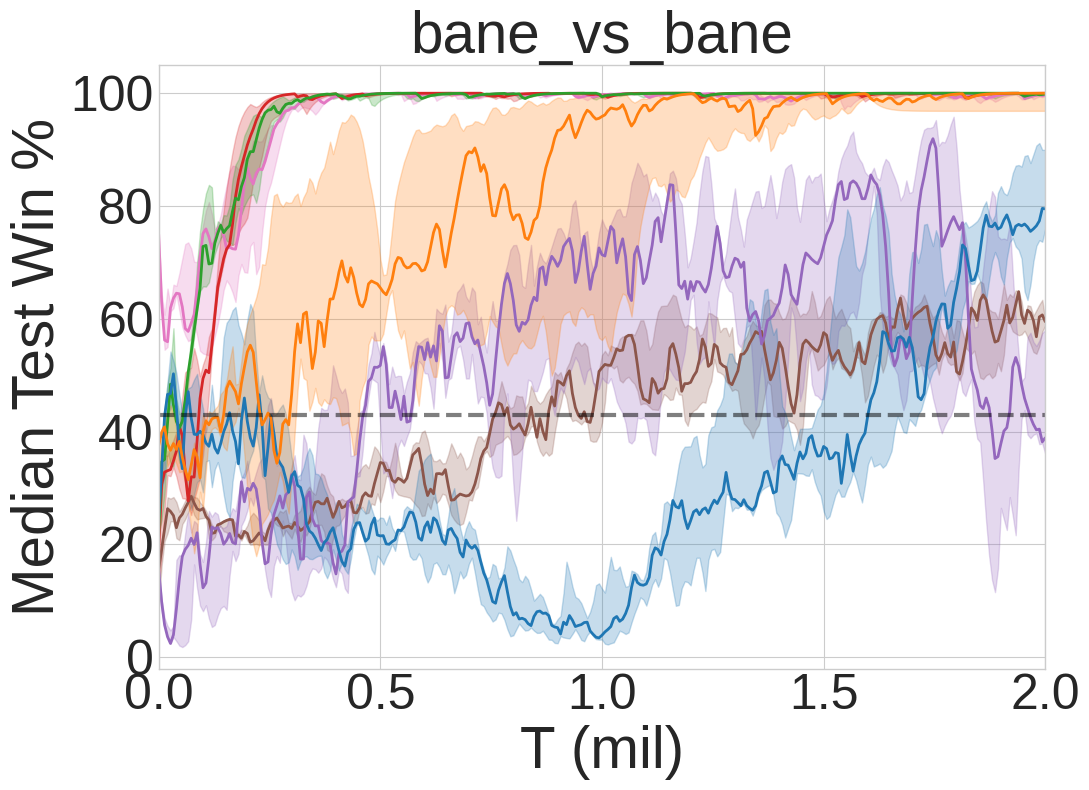}
	}
	\subfigure[{Super Hard}]{
		\label{fig:smac_mmm2}
		\includegraphics[width=0.3\textwidth]{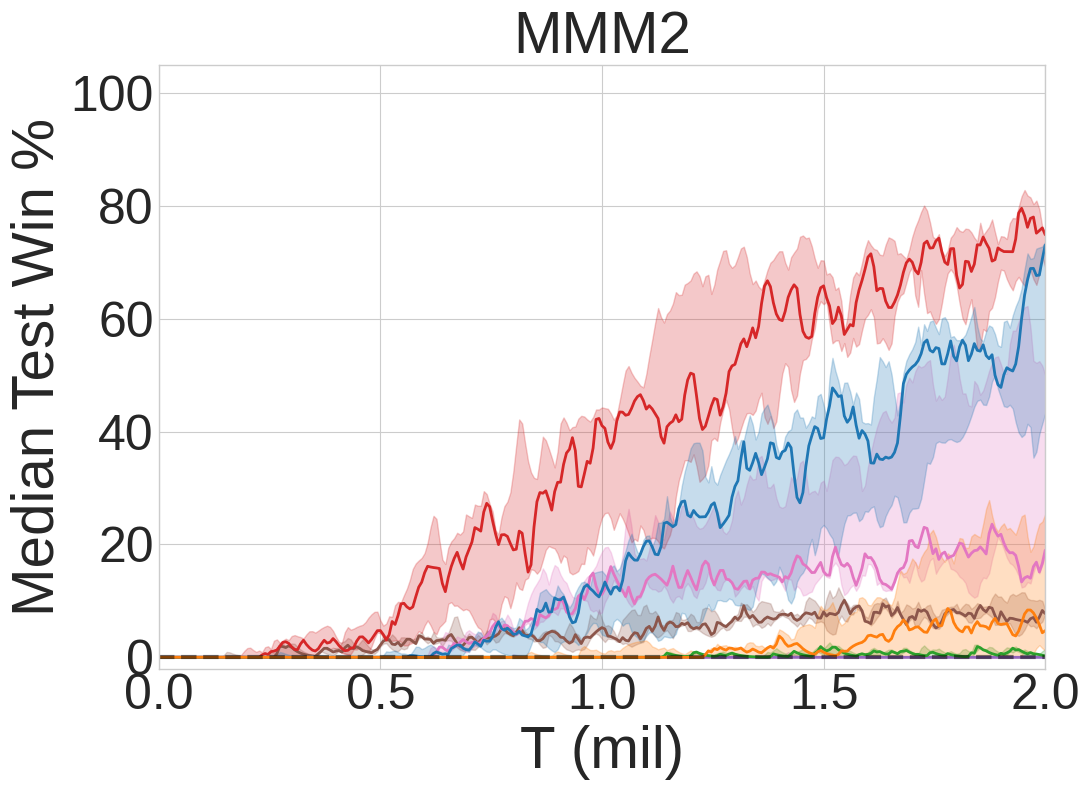}
	}
	\subfigure[{Super Hard}]{
		\label{fig:smac_27m_vs_30m}
		\includegraphics[width=0.3\textwidth]{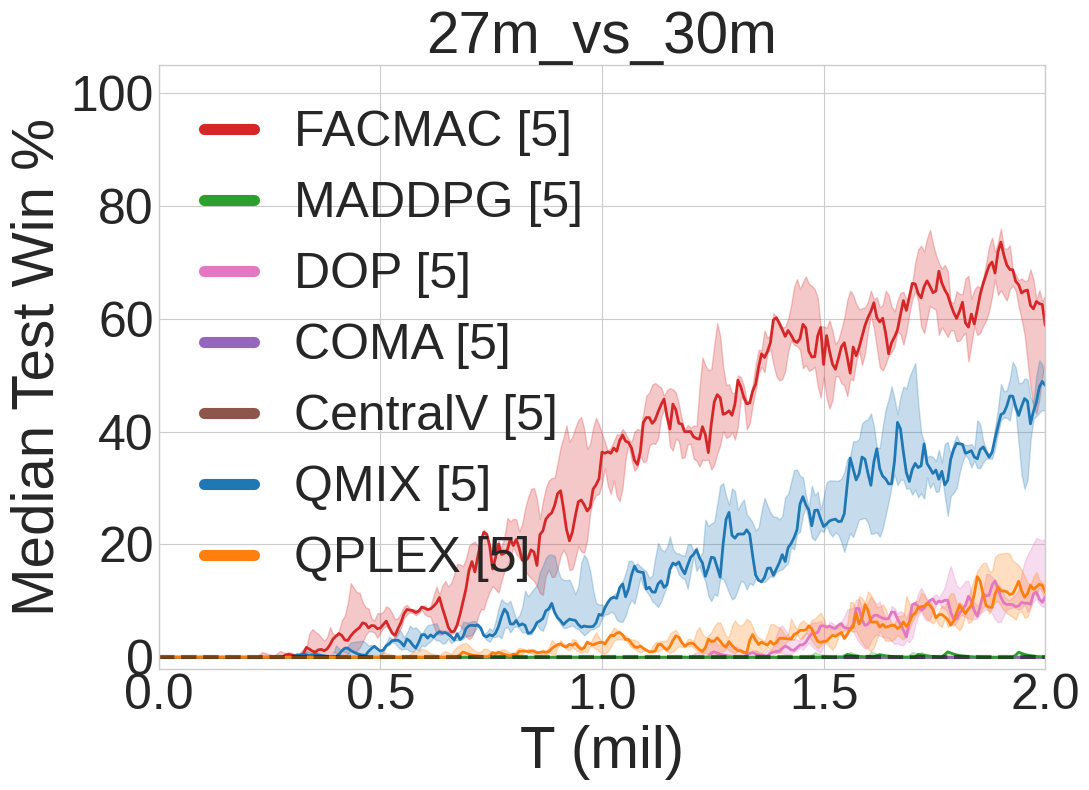}
	}
	\caption{Median test win \% on six different SMAC maps: (a) \textit{2s3z} (easy), (b) \textit{MMM} (easy),  (c) \textit{2c\_vs\_64zg} (hard), (d) \textit{bane\_vs\_bane} (hard), (e) \textit{MMM2} (super hard), and (f) \textit{27m\_vs\_30m} (super hard). The performance of the heuristic-based algorithm is shown as a dashed line.}
	\label{fig: smac_results}
\end{figure*}

\paragraph{\name{} outperforms MADDPG and other baselines in both discrete and continuous action tasks.} 
Figure \ref{fig: pp_results} and \ref{fig: mamujoco_results} illustrate the mean episode return attained by different methods on Continuous Predator-Prey with varying number of agents and different MAMuJoCo tasks, respectively. We can see that \name{} significantly outperforms MADDPG on all these continuous cooperative tasks, both in terms of absolute performance and learning speed.  On discrete SMAC tasks, Figure \ref{fig: smac_results} shows that \name{} performs significantly better than MADDPG on $4$ out of $6$ maps we tested, and achieves similar performance to MADDPG on the other $2$ maps. Additionally, on all $6$ SMAC maps, \name{} significantly outperforms all multi-agent actor-critic baselines (COMA, CentralV, and DOP), while DOP is recently claimed to be the first multi-agent actor-critic method that outperforms state-of-the-art valued-based methods on SMAC. \name{} is also competitive with state-of-the-art value-based methods (QMIX and QPLEX), with significantly better performance on \textit{MMM}, \textit{bane\_vs\_bane}, \textit{MMM2}, and \textit{27m\_vs\_30m}. These results demonstrate the benefits of our method for improving performance in  challenging cooperative tasks with discrete and continuous action spaces. 

In Continuous Predator-Prey, \name{}-vdn scales better than \name{} when the number of agents increases. However, on MAMuJoCo, \name{}-vdn performs drastically worse than \name{} in $2$-Agent Humanoid and ManyAgent Swimmer (with $10$ agents), demonstrating the necessity of the non-linear mixing of agent utilities and conditioning on the central state information in order to achieve competitive performance in such tasks. Furthermore, on SMAC, Figure \ref{fig: smac_results_diff_factorisations} in Appendix \ref{sec:nonmon_fac} shows that \name{} is noticeably more stable than \name{}-vdn and \name{}-vdn-s across different maps, and achieves significantly better performance on the \textit{super hard} map \textit{MMM2}. 

Interestingly, we find that \name{} performs similarly to COMIX on both Continuous Predator-Prey and MAMuJoCo tasks. As \name{} and COMIX use the same value factorisation as in QMIX and are both off-policy, this suggests that, in these continuous cooperative tasks, factorisation of the joint $Q$-value function plays a greater role in performance than the underlying algorithmic choices. On SMAC, however, \name{} performs significantly better than QMIX on \textit{MMM}, \textit{bane\_vs\_bane}, \textit{MMM2}, and \textit{27m\_vs\_30m}.
For instance, on \textit{bane\_vs\_bane}, a task with $24$ agents, while QMIX struggles to find the optimal policy with $2$ million timesteps, \name{}, with exactly the same value factorisation, can quickly recover the optimal policy and achieve $100\%$ test win rate. This shows the convergence advantages of policy gradient methods in this type of multi-agent settings \cite{wang2020towards}. 

\paragraph{\name{} scales better as the number of agents (and/or actions) and the complexity of the task increases.} As shown in Figure \ref{fig:pp_6a} and \ref{fig:pp_9a}, MADDPG performs poorly if we increase the number of agents in Continuous Predator-Prey, while both \name{} and \name{}-vdn achieve significantly better performance. The monolithic critic in MADDPG simply concatenates all agents' observations into a single input vector, which can be quite large when there are many agents and/or entities and make it more difficult to learn a good critic. Factoring the critic enables scalable critic learning, by combining individual agent utilities that condition on much smaller observations into a joint action-value function. 
Upon inspection of the learned policies, we find that, on both Continuous Predator-Prey tasks with $6$ agents and $9$ agents, \name{} agents learn to chase after the preys, while MADDPG agents quickly get stuck in sub-optimal coordination patterns -- learning to go after each other most of the time. On MAMuJoCo (shown in Figure \ref{fig: mamujoco_results}), similarly, the largest performance gap between \name{} and MADDPG can be seen on ManyAgent Swimmer (with $10$ agents), a task with the largest number of agents among three MAMuJoCo tasks tested.   

On SMAC (shown in Figure \ref{fig: smac_results}), the largest performance gap between \name{} and MADDPG can be seen on the challenging \textit{MMM2} and \textit{27m\_vs\_30m} with a large number of agents, which are classified as $2$ \textit{super hard} SMAC maps due to current methods' poor performance \cite{samvelyan19smac}. We can see that \name{} is able to learn to consistently defeat the enemy, whereas MADDPG fails to learn anything useful in both tasks. 
The second largest performance gap between \name{} and MADDPG can be seen on the hard map \textit{2c\_vs\_64zg}, where MADDPG not only performs significantly worse but also exhibits significantly more variance than \name{} across seeds. While there are only $2$ agents in this scenario, the number of actions each agent can choose is the largest among all $6$ maps tested as there are $64$ enemies. These results further demonstrate that \name{} scales better when the number of agents (and/or actions) and the complexity of the tasks increases.  

\paragraph{Factoring the critic can better take advantage of our centralised gradient estimator to optimise the agent policies when the number of agents and/or actions is large.}
We conduct ablation experiments to investigate the influence of factoring the critic and using the centralised gradient estimator in our method.
\name{} (without CPG) is our method without the centralised policy gradient. It uses a naive adaptation of the deterministic policy gradient used in MADDPG (shown in \eqref{eq:dpg2}). Thus, the only difference between \name{} (without CPG) and MADDPG is the previous one learns a non-linearly factored critic while the latter one learns a monolithic critic. We also evaluate MADDPG with our centralised policy gradient and refer to it as MADDPG (with CPG). 

\begin{figure}[t!]
   \centering
	\includegraphics[width=0.33\columnwidth]{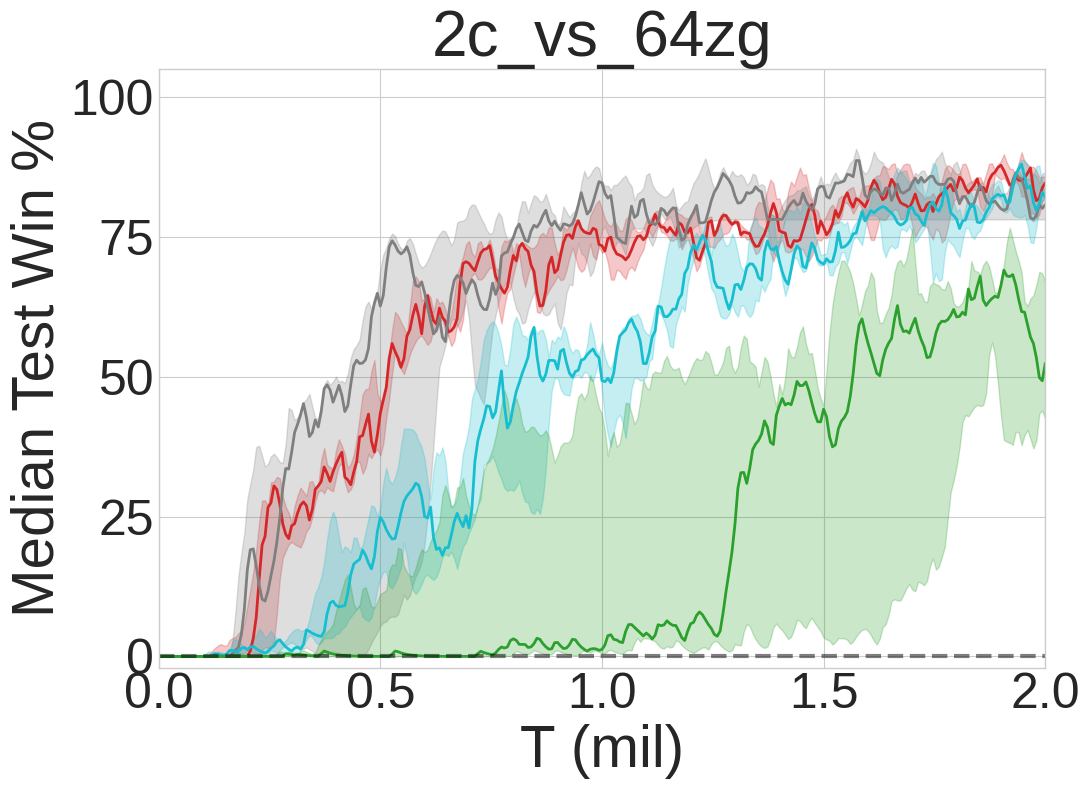}
	\includegraphics[width=0.33\columnwidth]{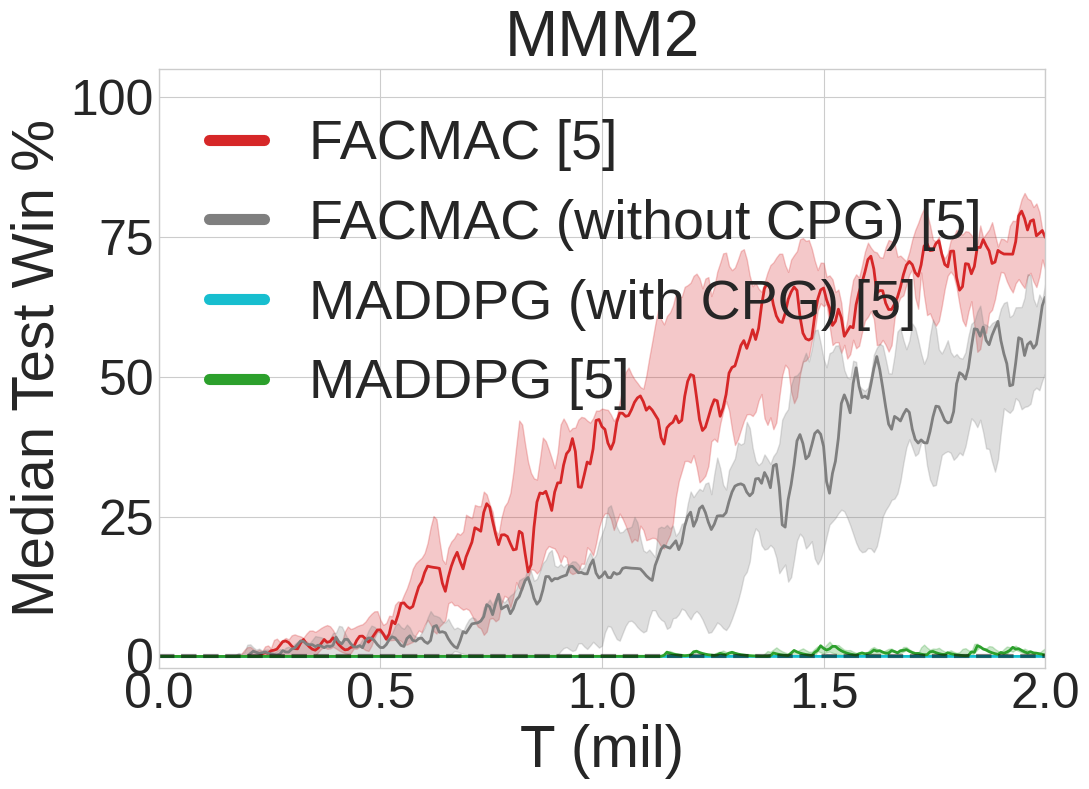}
    \includegraphics[width=0.33\columnwidth]{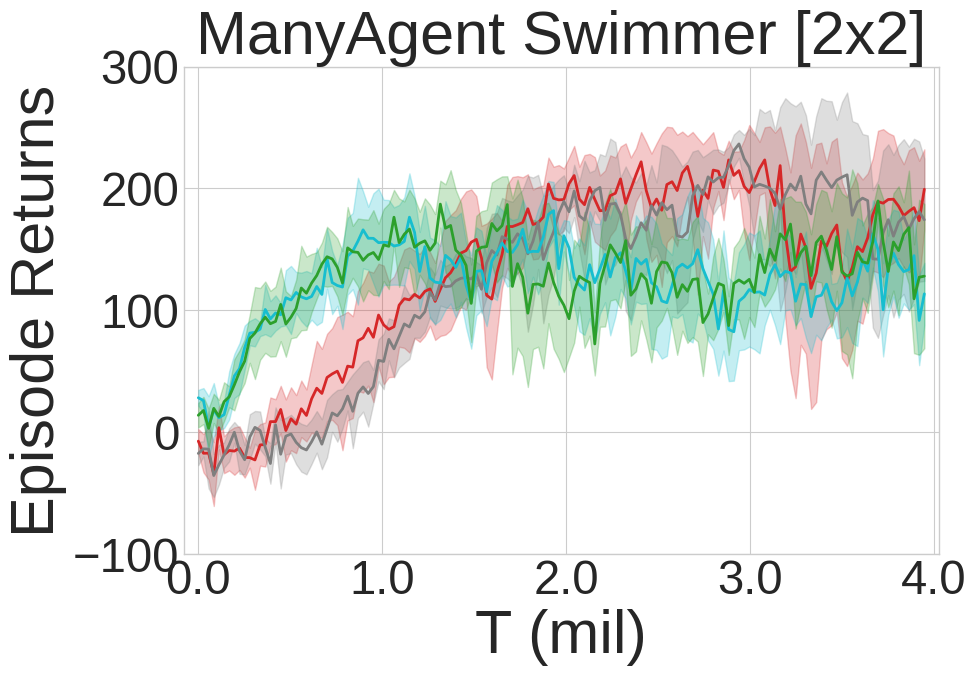}
	\includegraphics[width=0.33\columnwidth]{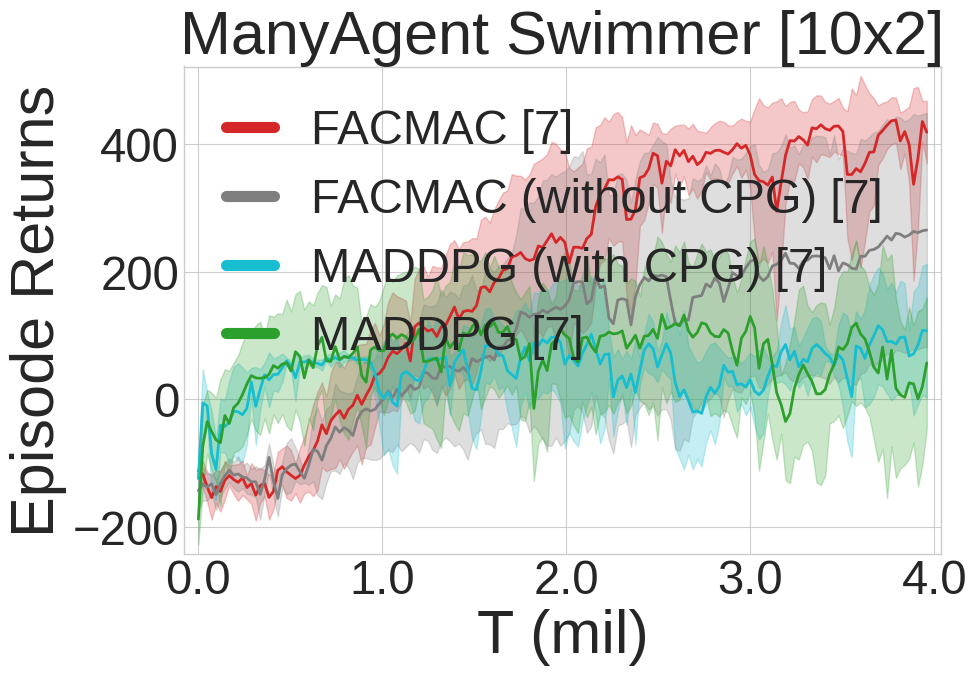}
    \caption{Ablations for different \name{} components on two SMAC maps (\textit{2c\_vs\_64zg} and \textit{MMM2}) and two MAMuJoCo tasks (ManyAgent Swimmer with $2$ and $10$ agents).}
    \label{fig:diff_pg}
\end{figure}

Figure \ref{fig:diff_pg} shows the results of these ablation experiments on SMAC and MAMuJoCo. We can see that \name{} (without CPG) performs significantly better than MADDPG on both SMAC maps tested, both in terms of absolute performance and learning speed, demonstrating the advantages of factoring the critic in challenging coordination problems. With the centralised policy gradient, MADDPG (with CPG) performs significantly better than MADDPG on \textit{2c\_vs\_64zg}. However, on the harder map \textit{MMM2}, MADDPG with both policy gradients fail to learn anything useful. By contrast, \name{} significantly outperforms \name{} (without CPG) on \textit{MMM2}, and has lower variance across seeds on \textit{2c\_vs\_64zg}. Furthermore, on ManyAgent Swimmer with $2$ agents, our centralised gradient estimator does not affect the performance of both MADDPG and \name{}. However, when the number of agents is increased to be $10$ in the same task, using the centralised gradient estimator can significantly improve the learning performance when learning a centralised but factored critic. 
These results demonstrate that factoring the critic can better take advantage of our centralised gradient estimator to optimise the agent policies when the number of agents and/or actions is large. 

\begin{figure}[t!]
\centering
    \includegraphics[width=0.33\columnwidth]{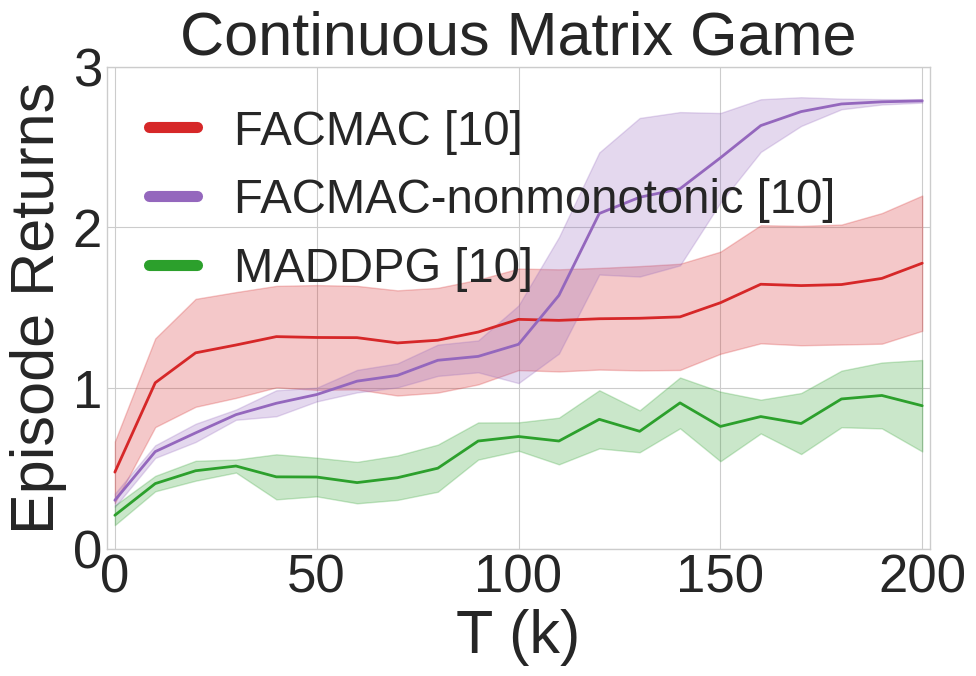}
    \includegraphics[width=0.32\columnwidth]{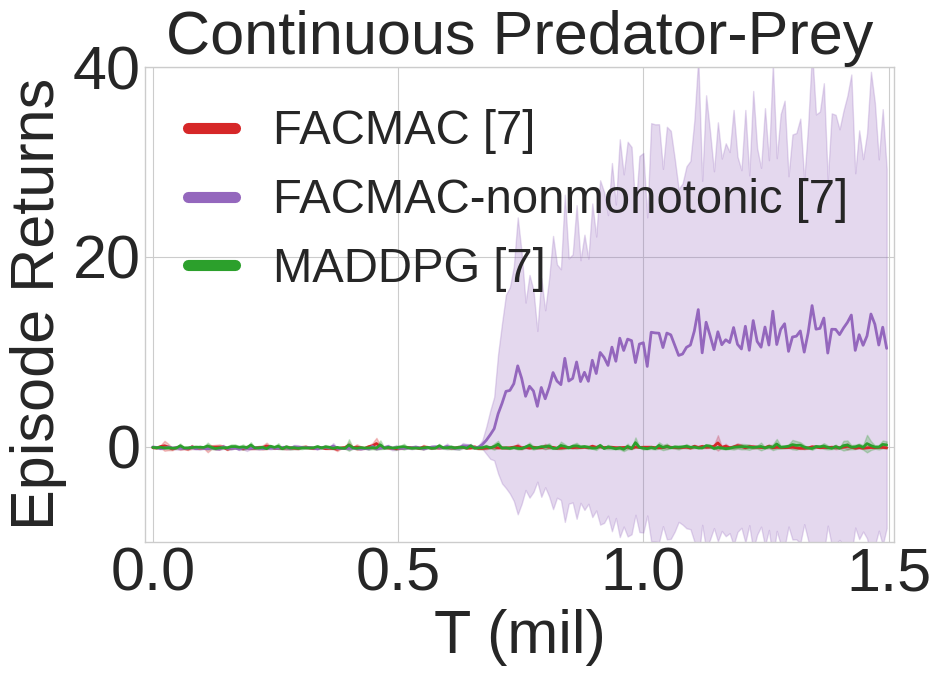}
    \caption{Mean episode return on (\textbf{Left}) Continuous Matrix Game and (\textbf{Right}) a variant of our Continuous Predator-Prey task (with $3$ agents and $1$ prey) with \textit{nonmonotonic} value functions.}
    \label{fig:cts_matrix_pp_nonmon}
\end{figure}

\paragraph{Nonmonotonically factored critics can solve tasks that cannot be solved with monolithic or monotonically factored critics.}  
In our multi-agent actor-critic framework, there are no inherent constraints on factoring the critic, we thus also employ a nonmonotonic factorisation and refer to it as \name{}-nonmonotonic (as discussed in Section \ref{sec: factored_critic}). As shown in Figure \ref{fig:cts_matrix_pp_nonmon} (left), on continuous matrix game (as discussed in Section \ref{sec: cpg}), \name{}-nonmonotonic can robustly learn the optimal policy, while both \name{} and MADDPG converge to some sub-optimal policy within $200k$ timesteps. On a variant of Continuous Predator-Prey with nonmonotonic value functions (see Appendix \ref{sec:pp_sup} for details about this task), Figure \ref{fig:cts_matrix_pp_nonmon} (right) shows that both \name{} and MADDPG fail to learn anything useful, while \name{}-nonmonotonic successfully learns to capture the prey. These results demonstrate that nonmonotonically factored critics can solve tasks that cannot be solved with monolithic or monotonically factored critics.

It is important to note that the relative performance of \name{} and \name{}-nonmonotonic is task dependent. On the original Continuous Predator-Prey task (with $3$ agents and $1$ prey), \name{}-nonmonotonic yields similar performance to \name{} (see Figure \ref{fig:cts_pp_nonmon} in Appendix \ref{sec:nonmon_fac}). On SMAC (see Figure \ref{fig: smac_results_diff_factorisations} in Appendix \ref{sec:nonmon_fac}), \name{}-nonmonotonic performs similarly to \name{} on easy maps, but exhibits significantly worse performance on harder maps. This shows that, in this type of tasks, using an unconstrained factored critic could lead to an increase in learning difficulty.
\section{Related Work}
\label{sec:related_work}
Value function factorisation \cite{koller_computing_1999} has been widely employed in value-based MARL algorithms.
VDN \cite{sunehag2018value} and QMIX \cite{rashid2018qmix} factor the joint action-value function into per-agent utilities that are combined via a simple summation or a monotonic mixing function respectively, to ensure consistency between the $\argmax$ of the centralised joint-action value function and the $\argmax$ of the decentralised polices. This monotonicity constraint, however, prevents them from representing joint action-value functions that are characterised as \textit{nonmonotonic} \cite{mahajan2019maven}, i.e., an agent's ordering over its own actions depends on other agents' actions.  
A large number of recent works \cite{son2019qtran,yang2020qatten,rashid2020weighted,wang2020qplex,son2020qtran++} thus focus on developing new value-based MARL algorithms that address this representational limitation, in order to learn a richer class of action-value functions. 

QTRAN \cite{son2019qtran} learns an
unrestricted joint action-value function and aims to solve a constrained optimisation problem in order to decentralise it, but has been shown to scale poorly to more complex tasks such as SMAC \cite{mahajan2019maven}. 
QPLEX \cite{wang2020qplex} takes advantage of the dueling network architecture to factor the joint action-value function in a manner that does not restrict the representational capacity, whilst also remaining easily decentralisable, but can still fail to solve simple tasks with nonmonotonic value functions \cite{rashid2020weighted}. Weighted QMIX \cite{rashid2020weighted} introduces a weighting scheme to place more importance on the better joint actions to learn a richer class of joint action-value functions. QTRAN++ \cite{son2020qtran++} addresses the gap between the empirical performance and theoretical guarantees of QTRAN. Our multi-agent actor-critic framework with decentralised actors and a centralised but factored critic, by contrast, provides a more direct and simpler way of coping with nonmonotonic tasks as one can simply factor the centralised critic in any manner without constraints. Additionally, our framework can be readily applied to tasks with continuous action spaces, whereas these value-based algorithms require additional algorithmic changes.  

Most state-of-the-art multi-agent actor-critic methods \cite{lowe2017multi,foerster_counterfactual_2018,iqbal2019actor,du_liir:_2019} learn a  centralised and monolithic critic conditioning on the global state and the joint action to stabilise learning. Even though the joint action-value function they can represent is not restricted, in practice they significantly underperform value-based methods like QMIX on the challenging SMAC benchmark \cite{rashid2020weighted,rashid2020monotonic}. In contrast, \name{} utilises a centralised but factored critic to allow it to scale to the more complex tasks in SMAC, and follows the centralised policy gradient instead of per-agent policy gradients. 

\citet{lyu2021contrasting} recently provide some interesting insights about the pros and cons of centralised and decentralised critics for on-policy actor-critic algorithms. One important issue that they highlight is that merely utilising a centralised critic does not necessarily lead to the learning of more coordinated behaviours. 
This is because the use of a per-agent policy gradient can lead to the agents getting stuck in sub-optimal solutions in which no one agents wishes to change their policy, as discussed in \ref{sec: cpg}. Our centralised policy gradient resolves this issue by taking full advantage of the centralised training paradigm to optimise over the joint-action policy, which allows us to reap the benefits of a centralised critic. Since \name{} is off-policy, we also benefit immensely from utilising a centralised critic over a decentralised one since we avoid the issues of non-stationarity when training on older data. 

\citet{zhou2020learning} propose to use a single centralised critic for MADDPG, whose weights are generated by hypernetworks that condition on the state, similarly to QMIX's mixing network without the monotonicity constraints. \name{} also uses a single centralised critic, but factorises it similarly to QMIX (not just using the mixing network) which allows for more efficient learning on more complex tasks. Of existing work, the deterministic decomposed policy gradients (DOP) algorithm proposed by \citet{wang2021dop} is perhaps most similar to our own approach. Deterministic DOP is off-policy and factors the centralised critic as a weighted linear sum of individual agent utilities and a state bias. It is limited to only considering linearly factored critics, which have limited representational capacity, whilst we are free to choose any method of factorisation in \name{} to allow for the learning of a richer class of action-value functions.
While they claim to be the first to introduce the idea of value function factorisation into the multi-agent actor-critic framework, it is actually first explored by \citet{bescuca_factorised_2019}, where a monotonically factored critic is learned for COMA \cite{foerster_counterfactual_2018}. However, their performance improvement on SMAC is limited since COMA requires on-policy learning and it is not straightforward to extend COMA to continuous action spaces.
Furthermore, both works only consider monotonically factored critics, whilst we employ a nonmonotonic factorisation and demonstrate its benefits.
We also investigate the benefits of learning a centralised but factored critic more thoroughly, providing a better understanding about the type of tasks that can benefit more from a factored critic. Furthermore, both deterministic DOP and LICA \cite{zhou2020learning} use a naive adaptation of the deterministic policy gradient used by MADDPG and suffer from the same problems as discussed in Section \ref{sec: cpg}, while our centralised policy gradients allow for better coordination across agents in certain tasks. 
\section{Conclusion}
This paper presented \name{}, a multi-agent actor-critic method that learns decentralised policies with a centralised but factored critic, working for both discrete and continuous cooperative tasks. We showed the advantages of both factoring the critic and using the new centralised gradient estimator in our approach. We also introduced a novel benchmark suite MAMuJoCo to demonstrate \name{}'s scalability to more complex continuous tasks. Our results on three different domains demonstrated \name{}'s superior performance over existing MARL algorithms. Future work will explore more forms of nonmonotonic factorisation to tackle tasks with nonmonotonic value functions.  

\bibliography{references}

\begin{thebibliography}{48}
\providecommand{\natexlab}[1]{#1}
\providecommand{\url}[1]{\texttt{#1}}
\expandafter\ifx\csname urlstyle\endcsname\relax
  \providecommand{\doi}[1]{doi: #1}\else
  \providecommand{\doi}{doi: \begingroup \urlstyle{rm}\Url}\fi

\bibitem[Ackermann et~al.(2019)Ackermann, Gabler, Osa, and
  Sugiyama]{ackermann_reducing_2019}
Johannes Ackermann, Volker Gabler, Takayuki Osa, and Masashi Sugiyama.
\newblock Reducing overestimation bias in multi-agent domains using double
  centralized critics.
\newblock \emph{arXiv preprint arXiv:1910.01465}, 2019.

\bibitem[Amos et~al.(2017)Amos, Xu, and Kolter]{amos_input_2016}
Brandon Amos, Lei Xu, and J~Zico Kolter.
\newblock Input convex neural networks.
\newblock In \emph{Proceedings of the 34th International Conference on Machine
  Learning-Volume 70}, pages 146--155. JMLR. org, 2017.

\bibitem[Bescuca(2019)]{bescuca_factorised_2019}
Marilena Bescuca.
\newblock Factorised critics in deep multi-agent reinforcement learning.
\newblock In \emph{Master Thesis, University of Oxford}, 2019.

\bibitem[Brockman et~al.(2016)Brockman, Cheung, Pettersson, Schneider,
  Schulman, Tang, and Zaremba]{brockman_openai_2016}
Greg Brockman, Vicki Cheung, Ludwig Pettersson, Jonas Schneider, John Schulman,
  Jie Tang, and Wojciech Zaremba.
\newblock Openai gym.
\newblock \emph{arXiv preprint arXiv:1606.01540}, 2016.

\bibitem[De~Boer et~al.(2005)De~Boer, Kroese, Mannor, and
  Rubinstein]{de_boer_tutorial_2005}
Pieter-Tjerk De~Boer, Dirk~P Kroese, Shie Mannor, and Reuven~Y Rubinstein.
\newblock A tutorial on the cross-entropy method.
\newblock \emph{Annals of operations research}, 134\penalty0 (1):\penalty0
  19--67, 2005.

\bibitem[Du et~al.(2019)Du, Han, Fang, Liu, Dai, and Tao]{du_liir:_2019}
Yali Du, Lei Han, Meng Fang, Ji~Liu, Tianhong Dai, and Dacheng Tao.
\newblock {LIIR}: Learning individual intrinsic reward in multi-agent
  reinforcement learning.
\newblock In \emph{Advances in Neural Information Processing Systems 32}, pages
  4405--4416, 2019.

\bibitem[Foerster et~al.(2018)Foerster, Farquhar, Afouras, Nardelli, and
  Whiteson]{foerster_counterfactual_2018}
Jakob Foerster, Gregory Farquhar, Triantafyllos Afouras, Nantas Nardelli, and
  Shimon Whiteson.
\newblock Counterfactual multi-agent policy gradients.
\newblock In \emph{Proceedings of the Thirty-Second AAAI Conference on
  Artificial Intelligence}, 2018.

\bibitem[Gu et~al.(2016)Gu, Lillicrap, Sutskever, and
  Levine]{gu_continuous_2016}
Shixiang Gu, Timothy Lillicrap, Ilya Sutskever, and Sergey Levine.
\newblock Continuous deep q-learning with model-based acceleration.
\newblock In \emph{International Conference on Machine Learning}, pages
  2829--2838, 2016.

\bibitem[Gupta et~al.(2017)Gupta, Egorov, and
  Kochenderfer]{gupta_cooperative_2017}
Jayesh~K Gupta, Maxim Egorov, and Mykel Kochenderfer.
\newblock Cooperative multi-agent control using deep reinforcement learning.
\newblock In \emph{International Conference on Autonomous Agents and Multiagent
  Systems}, pages 66--83. Springer, 2017.

\bibitem[Ha et~al.(2016)Ha, Dai, and Le]{ha_hypernetworks_2016}
David Ha, Andrew Dai, and Quoc~V Le.
\newblock Hypernetworks.
\newblock \emph{arXiv preprint arXiv:1609.09106}, 2016.

\bibitem[Iqbal and Sha(2019)]{iqbal2019actor}
Shariq Iqbal and Fei Sha.
\newblock Actor-attention-critic for multi-agent reinforcement learning.
\newblock In \emph{International Conference on Machine Learning}, pages
  2961--2970, 2019.

\bibitem[Jang et~al.(2016)Jang, Gu, and Poole]{jang2016categorical}
Eric Jang, Shixiang Gu, and Ben Poole.
\newblock Categorical reparameterization with gumbel-softmax.
\newblock \emph{arXiv preprint arXiv:1611.01144}, 2016.

\bibitem[Kalashnikov et~al.(2018)Kalashnikov, Irpan, Pastor, Ibarz, Herzog,
  Jang, Quillen, Holly, Kalakrishnan, Vanhoucke,
  et~al.]{kalashnikov_qt-opt:_2018}
Dmitry Kalashnikov, Alex Irpan, Peter Pastor, Julian Ibarz, Alexander Herzog,
  Eric Jang, Deirdre Quillen, Ethan Holly, Mrinal Kalakrishnan, Vincent
  Vanhoucke, et~al.
\newblock Qt-opt: Scalable deep reinforcement learning for vision-based robotic
  manipulation.
\newblock \emph{arXiv preprint arXiv:1806.10293}, 2018.

\bibitem[Kitano et~al.(1997)Kitano, Asada, Kuniyoshi, Noda, Osawa, and
  Matsubara]{kitano_robocup_1997}
Hiroaki Kitano, Minoru Asada, Yasuo Kuniyoshi, Itsuki Noda, Eiichi Osawa, and
  Hitoshi Matsubara.
\newblock Robocup: A challenge problem for ai.
\newblock \emph{AI magazine}, 18\penalty0 (1):\penalty0 73--73, 1997.

\bibitem[Koller and Parr(1999)]{koller_computing_1999}
Daphne Koller and Ronald Parr.
\newblock Computing factored value functions for policies in structured mdps.
\newblock In \emph{Proceedings of IJCAI}, pages 1332--1339, 1999.

\bibitem[Kraemer and Banerjee(2016)]{kraemer_multi-agent_2016}
Landon Kraemer and Bikramjit Banerjee.
\newblock Multi-agent reinforcement learning as a rehearsal for decentralized
  planning.
\newblock \emph{Neurocomputing}, 190:\penalty0 82--94, 2016.

\bibitem[Kurokawa et~al.(2008)Kurokawa, Tomita, Kamimura, Kokaji, Hasuo, and
  Murata]{kurokawa2008distributed}
Haruhisa Kurokawa, Kohji Tomita, Akiya Kamimura, Shigeru Kokaji, Takashi Hasuo,
  and Satoshi Murata.
\newblock Distributed self-reconfiguration of m-tran iii modular robotic
  system.
\newblock \emph{The International Journal of Robotics Research}, 27\penalty0
  (3-4):\penalty0 373--386, 2008.

\bibitem[Lillicrap et~al.(2016)Lillicrap, Hunt, Pritzel, Heess, Erez, Tassa,
  Silver, and Wierstra]{lillicrap_continuous_2015}
Timothy~P Lillicrap, Jonathan~J Hunt, Alexander Pritzel, Nicolas Heess, Tom
  Erez, Yuval Tassa, David Silver, and Daan Wierstra.
\newblock Continuous control with deep reinforcement learning.
\newblock In \emph{4th International Conference on Learning Representations,
  {ICLR}}, 2016.

\bibitem[Lin(1992)]{lin_self-improving_1992}
Long-Ji Lin.
\newblock Self-improving reactive agents based on reinforcement learning,
  planning and teaching.
\newblock \emph{Machine learning}, 8\penalty0 (3-4):\penalty0 293--321, 1992.

\bibitem[Liu et~al.(2019)Liu, Lever, Merel, Tunyasuvunakool, Heess, and
  Graepel]{liu_emergent_2019}
Siqi Liu, Guy Lever, Josh Merel, Saran Tunyasuvunakool, Nicolas Heess, and
  Thore Graepel.
\newblock Emergent coordination through competition.
\newblock \emph{arXiv preprint arXiv:1902.07151}, 2019.

\bibitem[Lowe et~al.(2017)Lowe, Wu, Tamar, Harb, Abbeel, and
  Mordatch]{lowe2017multi}
Ryan Lowe, Yi~Wu, Aviv Tamar, Jean Harb, OpenAI~Pieter Abbeel, and Igor
  Mordatch.
\newblock Multi-agent actor-critic for mixed cooperative-competitive
  environments.
\newblock In \emph{Advances in Neural Information Processing Systems}, pages
  6379--6390, 2017.

\bibitem[Lyu et~al.(2021)Lyu, Xiao, Daley, and Amato]{lyu2021contrasting}
Xueguang Lyu, Yuchen Xiao, Brett Daley, and Christopher Amato.
\newblock Contrasting centralized and decentralized critics in multi-agent
  reinforcement learning.
\newblock In \emph{Proceedings of the 20th International Conference on
  Autonomous Agents and Multi-Agent Systems}, 2021.

\bibitem[Mahajan et~al.(2019)Mahajan, Rashid, Samvelyan, and
  Whiteson]{mahajan2019maven}
Anuj Mahajan, Tabish Rashid, Mikayel Samvelyan, and Shimon Whiteson.
\newblock Maven: Multi-agent variational exploration.
\newblock In \emph{Advances in Neural Information Processing Systems}, pages
  7613--7624, 2019.

\bibitem[Mnih et~al.(2015)Mnih, Kavukcuoglu, Silver, Rusu, Veness, Bellemare,
  Graves, Riedmiller, Fidjeland, Ostrovski, et~al.]{mnih2015human}
Volodymyr Mnih, Koray Kavukcuoglu, David Silver, Andrei~A Rusu, Joel Veness,
  Marc~G Bellemare, Alex Graves, Martin Riedmiller, Andreas~K Fidjeland, Georg
  Ostrovski, et~al.
\newblock Human-level control through deep reinforcement learning.
\newblock \emph{nature}, 518\penalty0 (7540):\penalty0 529--533, 2015.

\bibitem[Nakagaki et~al.(2016)Nakagaki, Dementyev, Follmer, Paradiso, and
  Ishii]{nakagaki_chainform_2016}
Ken Nakagaki, Artem Dementyev, Sean Follmer, Joseph~A Paradiso, and Hiroshi
  Ishii.
\newblock Chainform: A linear integrated modular hardware system for shape
  changing interfaces.
\newblock In \emph{Proceedings of the 29th Annual Symposium on User Interface
  Software and Technology}, pages 87--96, 2016.

\bibitem[Oliehoek et~al.(2008)Oliehoek, Spaan, and {Nikos
  Vlassis}]{oliehoek_optimal_2008}
Frans~A. Oliehoek, Matthijs T.~J. Spaan, and {Nikos Vlassis}.
\newblock Optimal and approximate {Q}-value functions for decentralized pomdps.
\newblock \emph{JAIR}, 32:\penalty0 289--353, 2008.

\bibitem[Oliehoek et~al.(2016)Oliehoek, Amato, et~al.]{oliehoek_concise_2016}
Frans~A Oliehoek, Christopher Amato, et~al.
\newblock \emph{A concise introduction to decentralized POMDPs}, volume~1.
\newblock Springer, 2016.

\bibitem[OpenAI(2020)]{openai_openaibaselines_2020}
OpenAI.
\newblock openai/baselines, May 2020.
\newblock original-date: 2017-05-24T01:58:13Z.

\bibitem[Pan et~al.(2021)Pan, Rashid, Peng, Huang, and
  Whiteson]{pan2021softmax}
Ling Pan, Tabish Rashid, Bei Peng, Longbo Huang, and Shimon Whiteson.
\newblock Softmax with regularization: Better value estimation in multi-agent
  reinforcement learning.
\newblock \emph{arXiv preprint arXiv:2103.11883}, 2021.

\bibitem[Rashid et~al.(2018)Rashid, Samvelyan, Witt, Farquhar, Foerster, and
  Whiteson]{rashid2018qmix}
Tabish Rashid, Mikayel Samvelyan, Christian~Schroeder Witt, Gregory Farquhar,
  Jakob Foerster, and Shimon Whiteson.
\newblock Qmix: Monotonic value function factorisation for deep multi-agent
  reinforcement learning.
\newblock In \emph{International Conference on Machine Learning}, pages
  4292--4301, 2018.

\bibitem[Rashid et~al.(2020{\natexlab{a}})Rashid, Farquhar, Peng, and
  Whiteson]{rashid2020weighted}
Tabish Rashid, Gregory Farquhar, Bei Peng, and Shimon Whiteson.
\newblock Weighted qmix: Expanding monotonic value function factorisation.
\newblock In \emph{Advances in neural information processing systems},
  2020{\natexlab{a}}.

\bibitem[Rashid et~al.(2020{\natexlab{b}})Rashid, Samvelyan, De~Witt, Farquhar,
  Foerster, and Whiteson]{rashid2020monotonic}
Tabish Rashid, Mikayel Samvelyan, Christian~Schroeder De~Witt, Gregory
  Farquhar, Jakob Foerster, and Shimon Whiteson.
\newblock Monotonic value function factorisation for deep multi-agent
  reinforcement learning.
\newblock \emph{JMLR}, 21:\penalty0 178:1--178:51, 2020{\natexlab{b}}.

\bibitem[Riedmiller et~al.(2009)Riedmiller, Gabel, Hafner, and
  Lange]{riedmiller_reinforcement_2009}
Martin Riedmiller, Thomas Gabel, Roland Hafner, and Sascha Lange.
\newblock Reinforcement learning for robot soccer.
\newblock \emph{Autonomous Robots}, 27\penalty0 (1):\penalty0 55--73, 2009.

\bibitem[Samvelyan et~al.(2019)Samvelyan, Rashid, de~Witt, Farquhar, Nardelli,
  Rudner, Hung, Torr, Foerster, and Whiteson]{samvelyan19smac}
Mikayel Samvelyan, Tabish Rashid, Christian~Schroeder de~Witt, Gregory
  Farquhar, Nantas Nardelli, Tim G.~J. Rudner, Chia-Man Hung, Philiph H.~S.
  Torr, Jakob Foerster, and Shimon Whiteson.
\newblock {The} {StarCraft} {Multi}-{Agent} {Challenge}.
\newblock \emph{CoRR}, abs/1902.04043, 2019.

\bibitem[Son et~al.(2019)Son, Kim, Kang, Hostallero, and Yi]{son2019qtran}
Kyunghwan Son, Daewoo Kim, Wan~Ju Kang, David~Earl Hostallero, and Yung Yi.
\newblock Qtran: Learning to factorize with transformation for cooperative
  multi-agent reinforcement learning.
\newblock \emph{arXiv preprint arXiv:1905.05408}, 2019.

\bibitem[Son et~al.(2020)Son, Ahn, Reyes, Shin, and Yi]{son2020qtran++}
Kyunghwan Son, Sungsoo Ahn, Roben~Delos Reyes, Jinwoo Shin, and Yung Yi.
\newblock Qtran++: Improved value transformation for cooperative multi-agent
  reinforcement learning.
\newblock \emph{arXiv preprint arXiv:2006.12010}, 2020.

\bibitem[Stone and Sutton(2001)]{stone_scaling_2001}
Peter Stone and Richard~S. Sutton.
\newblock Scaling reinforcement learning toward {RoboCup} soccer.
\newblock In \emph{Icml}, volume~1, pages 537--544. Citeseer, 2001.

\bibitem[Sunehag et~al.(2018)Sunehag, Lever, Gruslys, Czarnecki, Zambaldi,
  Jaderberg, Lanctot, Sonnerat, Leibo, Tuyls, et~al.]{sunehag2018value}
Peter Sunehag, Guy Lever, Audrunas Gruslys, Wojciech~Marian Czarnecki,
  Vin{\'\i}cius~Flores Zambaldi, Max Jaderberg, Marc Lanctot, Nicolas Sonnerat,
  Joel~Z Leibo, Karl Tuyls, et~al.
\newblock Value-decomposition networks for cooperative multi-agent learning
  based on team reward.
\newblock In \emph{AAMAS}, pages 2085--2087, 2018.

\bibitem[Todorov et~al.(2012)Todorov, Erez, and Tassa]{todorov_mujoco:_2012}
Emanuel Todorov, Tom Erez, and Yuval Tassa.
\newblock Mujoco: A physics engine for model-based control.
\newblock In \emph{2012 IEEE/RSJ International Conference on Intelligent Robots
  and Systems}, pages 5026--5033, 2012.

\bibitem[Wang et~al.(2020{\natexlab{a}})Wang, Ren, Han, and
  Zhang]{wang2020towards}
Jianhao Wang, Zhizhou Ren, Beining Han, and Chongjie Zhang.
\newblock Towards understanding linear value decomposition in cooperative
  multi-agent q-learning.
\newblock \emph{arXiv preprint arXiv:2006.00587}, 2020{\natexlab{a}}.

\bibitem[Wang et~al.(2020{\natexlab{b}})Wang, Ren, Liu, Yu, and
  Zhang]{wang2020qplex}
Jianhao Wang, Zhizhou Ren, Terry Liu, Yang Yu, and Chongjie Zhang.
\newblock Qplex: Duplex dueling multi-agent q-learning.
\newblock \emph{arXiv preprint arXiv:2008.01062}, 2020{\natexlab{b}}.

\bibitem[Wang et~al.(2018)Wang, Liao, Ba, and Fidler]{wang_nervenet_2018}
Tingwu Wang, Renjie Liao, Jimmy Ba, and Sanja Fidler.
\newblock {NerveNet}: learning structured policy with graph neural networks.
\newblock In \emph{6th International Conference on Learning Representations,
  {ICLR}}, 2018.

\bibitem[Wang et~al.(2021)Wang, Han, Wang, Dong, and Zhang]{wang2021dop}
Yihan Wang, Beining Han, Tonghan Wang, Heng Dong, and Chongjie Zhang.
\newblock Dop: Off-policy multi-agent decomposed policy gradients.
\newblock In \emph{9th International Conference on Learning Representations,
  {ICLR}}, 2021.

\bibitem[Wei and Luke(2016)]{wei2016lenient}
Ermo Wei and Sean Luke.
\newblock Lenient learning in independent-learner stochastic cooperative games.
\newblock \emph{The Journal of Machine Learning Research}, 17\penalty0
  (1):\penalty0 2914--2955, 2016.

\bibitem[Wright et~al.(2012)Wright, Buchan, Brown, Geist, Schwerin, Rollinson,
  Tesch, and Choset]{wright2012design}
Cornell Wright, Austin Buchan, Ben Brown, Jason Geist, Michael Schwerin, David
  Rollinson, Matthew Tesch, and Howie Choset.
\newblock Design and architecture of the unified modular snake robot.
\newblock In \emph{2012 IEEE International Conference on Robotics and
  Automation}, pages 4347--4354. IEEE, 2012.

\bibitem[Yang et~al.(2020)Yang, Hao, Liao, Shao, Chen, Liu, and
  Tang]{yang2020qatten}
Yaodong Yang, Jianye Hao, Ben Liao, Kun Shao, Guangyong Chen, Wulong Liu, and
  Hongyao Tang.
\newblock Qatten: A general framework for cooperative multiagent reinforcement
  learning.
\newblock \emph{arXiv preprint arXiv:2002.03939}, 2020.

\bibitem[Yim et~al.(2002)Yim, Zhang, and Duff]{yim2002modular}
Mark Yim, Ying Zhang, and David Duff.
\newblock Modular robots.
\newblock \emph{IEEE Spectrum}, 39\penalty0 (2):\penalty0 30--34, 2002.

\bibitem[Zhou et~al.(2020)Zhou, Liu, Sui, Li, and Chung]{zhou2020learning}
Meng Zhou, Ziyu Liu, Pengwei Sui, Yixuan Li, and Yuk~Ying Chung.
\newblock Learning implicit credit assignment for multi-agent actor-critic.
\newblock In \emph{Advances in neural information processing systems}, 2020.

\end{thebibliography}
\bibliographystyle{plainnat}
\onecolumn
\newpage
\appendix
\section{Multi-Agent MuJoCo}
\label{sec:mamujoco_sup}
While several MARL benchmarks with continuous action spaces have been released, few are simultaneously diverse, fully cooperative, decentralisable, and admit partial observability. The Multi-Agent Particle suite \citep{lowe2017multi} features a few decentralisable tasks in a fully observable planar point mass toy environment. Presumably due to its focus on real-world robotic control, RoboCup Soccer Simulation \citep{kitano_robocup_1997,stone_scaling_2001,riedmiller_reinforcement_2009} does not currently feature an easily configurable software interface for MARL, nor suitable AI-controlled benchmark opponents. \citet{liu_emergent_2019} introduce MuJoCo Soccer Environment, a multi-agent soccer environment with continuous simulated physics that cannot be used in a purely cooperative setting and does not admit partial observability. 

To demonstrate \name{}'s scalability to more complex continuous domains and to stimulate more progress in continuous MARL, we develop \textit{Multi-Agent MuJoCo} (MAMuJoCo), a novel benchmark for continuous cooperative multi-agent robotic control. 
Starting from the popular fully observable single-agent robotic MuJoCo \citep{todorov_mujoco:_2012} control suite included with OpenAI Gym \citep{brockman_openai_2016}, we create a wide variety of novel scenarios in which multiple agents within a single robot have to solve a task cooperatively.
Single-robot multi-agent tasks in MAMuJoCo arise by first representing a given single robotic agent as a \textit{body graph}, where vertices (joints) are connected by adjacent edges (body segments), as shown in Figure \ref{fig:multiagent_mujoco}.
We then partition the body graph into disjoint sub-graphs, one for each agent, each of which contains one or more joints that can be controlled. Note that in ManyAgent Swimmer (see Figure \ref{fig:multiagent_mujoco}A) and ManyAgent Ant (see  Figure \ref{fig:multiagent_mujoco}K), the number of agents are not limited by the given single robotic agent. 

Each agent's action space in MAMuJoCo is given by the joint action space over all motors controllable by that agent. For example, the agent corresponding to the green partition in 2-Agent HalfCheetah (Figure \ref{fig:multiagent_mujoco}C) %
consists of three joints (joint ids 1, 2, and 3) and four adjacent body segments. 
Each joint has an action space $[-1,1]$, so the action space for each agent is a 3-dimensional vector with each entry in $[-1,1]$.

\begin{figure}[ht!]
	\centerline{\includegraphics[width=0.45\textwidth]{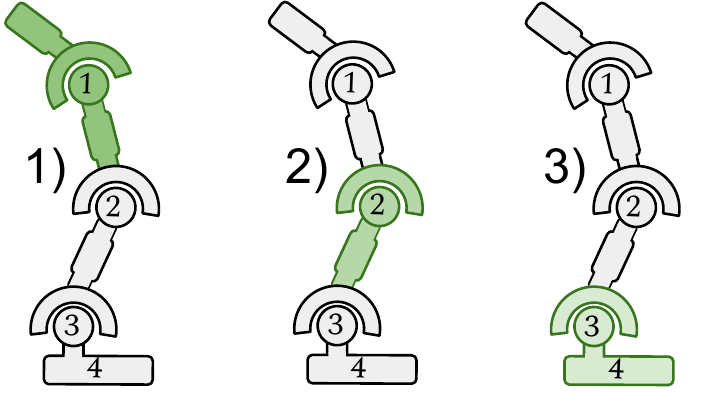}}
	\caption{\textbf{Observations by distance for 3-Agent Hopper (as seen from agent 1).} Each corresponds to joints and body parts observable at 1) zero graph distance from agent 1, 2) one unit graph distance from agent 1, and 3) two unit graph distances from agent 1.}
	\label{fig:multiagent_mujoco_obs}
\end{figure}

For each agent $a$, observations are constructed in a two-stage process. First, we infer which body segments and joints are observable by agent $a$. Each agent can always observe all joints within its own sub-graph. A configurable parameter $k\geq 0$ determines the maximum graph distance to the agent's subgraph at which joints are observable (see Figure \ref{fig:multiagent_mujoco_obs} for an example). Body segments directly attached to observable joints are themselves observable. The agent observation is then given by a fixed order concatenation of the representation vector of each observable graph element. Depending on the environment and configuration, representation vectors may include attributes such as position, velocity, and external body forces. In addition to joint and body-segment specific observation categories, agents can also be configured to observe the position and/or velocity attributes of the robot's central torso.

Restricting both the observation distance $k$, as well as limiting the set of observable element categories imposes partial observability. However, task goals remain unchanged from the single-agent variants, except that the goals must be reached collaboratively by multiple agents: we simply repurpose the original single-agent reward signal as a team reward signal. Default team reward is summarised in Table \ref{tab:multiagent_mujoco_goals}.

The most similar existing environments, though not as diverse as MAMuJoCo, are the decomposed MuJoCo environments Centipede and Snakes \citep{wang_nervenet_2018}. The latter is similar to MAMuJoCo's 2-Agent Swimmer. \citet{ackermann_reducing_2019} evaluate on one environment similar to a configuration of 2-Agent Ant, but, similarly to \citet{gupta_cooperative_2017}, do not consider tasks across different numbers of agents and MuJoCo scenarios. 

\begin{table*}[hbt!]
\scriptsize
\begin{center}
 \begin{tabular}{||c c c c||} 
 \hline
 \textbf{Task} & \textbf{Goal} & \textbf{Special observations} & \textbf{Reward function} \\ [0.5ex] 
 \hline\hline
 2-Agent Swimmer &  Maximise $+$ve $x$-speed.  & \makecell{All agents can observe \\velocities of the central torso.} & $\frac{\Delta x}{\Delta t} + 0.0001\alpha$  \\ 
 2-Agent Reacher & \makecell{Fingertip (green) needs\\ to reach target at random\\ location (red).} & \makecell{Target is only \\visible to green agent.} & \makecell{$-\left\lVert\text{distance from fingertip to target}\right\rVert^2_2$\\$+\alpha$}\\
 2-Agent Ant & Maximise $+$ve $x$-speed.  & \makecell{All agents can observe \\velocities of the central torso.} & \makecell{$\frac{\Delta x}{\Delta t} + 5\cdot10^{-4}\left\lVert\text{external contact forces}\right\rVert^2_2$ \\ $+ 0.5\alpha + 1$} \\
 2-Agent Ant Diag & Maximise $+$ve $x$-speed.  & \makecell{All agents can observe \\velocities of the central torso.} & \makecell{$\frac{\Delta x}{\Delta t} + 5\cdot10^{-4}\left\lVert\text{external contact forces}\right\rVert^2_2$ \\ $+ 0.5\alpha + 1$} \\
 2-Agent HalfCheetah & Maximise $+$ve $x$-speed.  & -  & $\frac{\Delta x}{\Delta t} + 0.1\alpha$ \\
 2-Agent Humanoid & Maximise $+$ve $x$-speed.   & - &  \makecell{$0.25\frac{\Delta x}{\Delta t} + \min(10, $\\$5\cdot10^{-6}\left\lVert\text{external contact forces}\right\rVert^2_2)$} \\
 \makecell{2-Agent\\ HumanoidStandup} & Maximise $+$ve $x$-speed. & - & \makecell{$\frac{y}{\Delta t} + \min(10, $\\$5\cdot10^{-6}\left\lVert\text{external contact forces}\right\rVert^2_2)$} \\ 
 3-Agent Hopper &  Maximise $+$ve $x$-speed.   & - & $\frac{\Delta x}{\Delta t} + 0.001\alpha + 1.0$ \\
 4-Agent Ant & Maximise $+$ve $x$-speed.  & \makecell{All agents can observe \\velocities of the central torso.} & \makecell{$\frac{\Delta x}{\Delta t} + 5\cdot10^{-4}\left\lVert\text{external contact forces}\right\rVert^2_2$ \\ $+ 0.5\alpha + 1$} \\
 6-Agent HalfCheetah & Maximise $+$ve $x$-speed. & - & \makecell{$0.25\frac{\Delta x}{\Delta t} + \min(10, $\\$5\cdot10^{-6}\left\lVert\text{external contact forces}\right\rVert^2_2)$}  \\
 ManyAgent Swimmer &  Maximise $+$ve $x$-speed.  & \makecell{All agents can observe \\velocities of the central torso.} & $\frac{\Delta x}{\Delta t} + 0.0001\alpha$  \\ 
 ManyAgent Ant & Maximise $+$ve $x$-speed.  & \makecell{All agents can observe \\velocities of the central torso.} & \makecell{$\frac{\Delta x}{\Delta t} + 5\cdot10^{-4}\left\lVert\text{external contact forces}\right\rVert^2_2$ \\ $+ 0.5\alpha + 1$} \\
 \hline
 \end{tabular}
 \end{center}
\caption{Overview of tasks contained in MAMuJoCo. We define $\alpha$ as an action regularisation term $-\left\lVert\mathbf{u}\right\rVert^2_2$.}
\label{tab:multiagent_mujoco_goals}
\end{table*}

\section{COVDN and COMIX}
\label{sec:comix}
$Q$-learning has shown considerable success in multi-agent settings with discrete action spaces \citep{sunehag2018value,rashid2018qmix,son2019qtran,rashid2020weighted,wang2020qplex}. However, performing greedy action selection in $Q$-learning requires evaluating $\mathop{\arg\max}_\mathbf{u} Q_{tot}(\boldsymbol{\tau}, \mathbf{u}, s)$, where $Q_{tot}$ is the joint action-value function. In discrete action spaces, this operation can be performed efficiently through enumeration (unless the action space is extremely large). In continuous action spaces, however, enumeration is impossible. Hence, existing continuous $Q$-learning approaches in single-agent settings either impose constraints on the form of $Q$-value to make maximisation easy \citep{gu_continuous_2016,amos_input_2016}, at the expense of estimation bias, or perform only approximate greedy action selection \citep{kalashnikov_qt-opt:_2018}. 
Neither approach scales easily to the large joint action spaces inherent to multi-agent settings, as 1) the joint action space grows exponentially in the number of agents, and 2) training $Q_{tot}$ required for greedy action selection becomes impractical when there are many agents. 

This highlights the importance of learning a centralised but factored $Q_{tot}$. To factor large joint action spaces efficiently in a decentralisable fashion, COVDN represents the joint action-value function $Q_{tot}$ as a sum of the per-agent utilities $Q_a$ as in VDN \cite{sunehag2018value}, while COMIX represents $Q_{tot}$ as a non-linear monotonic combination of $Q_a$ as in QMIX \citep{rashid2018qmix}.
COVDN and COMIX are thus simple variants of VDN and QMIX, respectively, that scale to continuous action spaces. 
They both perform approximate greedy selection of actions $u_a$ with respect to utility functions $Q_a$ for each agent $a$ using the cross-entropy method (CEM) \citep{de_boer_tutorial_2005}.
CEM is a sampling-based derivative-free heuristic search method that has been successfully used to find approximate maxima of nonconvex $Q$-networks in a number of single-agent robotic control tasks \citep{kalashnikov_qt-opt:_2018}. 
The centralised but factored $Q_{tot}$ allows us to use CEM to sample actions for each agent independently and to use the individual utility function $Q_a$ to guide the selection of maximal actions. 
Algorithm \ref{alg:cem} outlines the full CEM process used in both COVDN and COMIX.
Algorithm \ref{alg:comix} outlines the full process for COMIX. 
Note we do not consider COVDN and COMIX significant algorithmic contributions but instead merely baseline algorithms.

\begin{algorithm}[ht!]
\begin{algorithmic}
	\FOR{$a:=1,\ a\leq N$}
		\STATE ${\boldsymbol\mu}_a\leftarrow \mathbf{0}\in\mathbb{R}^{|\mathcal{A}_a|}$
		\STATE ${\boldsymbol\sigma}_a\leftarrow \mathbf{1}\in\mathbb{R}^{|\mathcal{A}_a|}$
		\FOR{$i:=1,\ i\leq n_c$}		
			\FOR{$j:=1,\ j\leq d_i$}
				\STATE $\mathbf{v'}_{aj}\sim \mathcal{N}({\boldsymbol\mu}_a, {\boldsymbol\sigma}_a) $		
				\STATE $\mathbf{v}_{aj} \leftarrow \tanh(\mathbf{v'}_{aj})$					\STATE $q_{aj} \leftarrow Q_a(\tau_a,\mathbf{v}_{aj})$
				\STATE $j\leftarrow j+1$
			\ENDFOR		
			\IF{$i<n_c$}
				\STATE \text{\scriptsize $U\leftarrow \left\{\mathbf{v'}_{al}\ |\ q_{al} \in \text{top}k_i(q_{a1},\dots, q_{ad_i}), \forall l\in\left\{1\dots N\right\}\right\}$} 				
				\STATE ${\boldsymbol\mu_a}\leftarrow\text{sample\_mean}(U)$
				\STATE ${\boldsymbol\sigma_a}\leftarrow\text{sample\_std}(U)$
				
			\ELSE

				\STATE $m\leftarrow \mathop{\arg\max}_j q_{aj}$
				
				\STATE $\mathbf{u}_a\leftarrow \mathbf{v}_{am}$
				
			\ENDIF
			\STATE $i\leftarrow i+1$		
		\ENDFOR
		
		\STATE $a\leftarrow a+1$
	
	\ENDFOR
	\STATE \textbf{return\ } $\left\langle\mathbf{u}_1,\dots,\mathbf{u}_{n}\right\rangle$
\end{algorithmic}
\caption{For each agent $a$, we perform $n_c$ CEM iterations. Hyper-parameters $d_i\in\mathbb{N}$ control how many actions are sampled at the $i$th iteration.} 
\label{alg:cem}
\end{algorithm}

\begin{algorithm}[ht!]
\begin{algorithmic}
	\STATE $\text{Initialise ReplayBuffer}, \theta, \theta^-, \phi, \phi^-$
	\FOR{each training episode $e$}
		\STATE $s_0, \mathbf{z}_0 \leftarrow \text{EnvInit}()$
		\FOR{$t:=0$ until $t=T$ step $1$}
			\STATE $\mathbf{u}_t\leftarrow \text{CEM}(Q_1,\dots,Q_N, \tau^1_t,\dots,\tau^N_t)$
			\STATE $\langle s_{t+1}, \mathbf{z}_{t+1}, r_t \rangle \leftarrow \text{EnvStep}(\mathbf{u}_t)$	
			\vspace{.5mm}
			\STATE  $\text{ReplayBuffer}\leftarrow \left\langle s_t,\mathbf{u}_t,
						\mathbf{z}_t,r_t,s_{t+1},\vec z_{t+1}\right\rangle$				
		\ENDFOR
		\STATE $\{\langle s_i, \vec u_i, \vec z_i, r_i, s'_i,\vec z'_i \rangle\}_{i=1}^b
					\sim \text{ReplayBuffer}$
		\vspace{1mm}
		\STATE $y_i \leftarrow r_i
					+ \gamma\mathop{\max}\limits_{\mathbf{u}'_i}
				Q_{\text{tot}}(s'_i,\vec z'_i,\vec u'_i;\vec\theta^-\!\!\!,\phi^-),
				\;\forall i$
		\vspace{-2.25mm}
		\STATE $\mathcal{L} \leftarrow
				\sum\limits_{i=1}^{b}\!
				\Big(y_i - 
					Q_{\text{tot}}(s_i, \vec z_i,\vec u_i; \vec\theta, \phi)
				\Big)^{\!2}$
		\vspace{.5mm}
		\STATE $\vec\theta \leftarrow \vec\theta 
				- \alpha \, \nabla_{\!\vec\theta} \Set L$
		\STATE $\phi \leftarrow \phi - \alpha \, \nabla_{\!\phi} \Set L$
	\ENDFOR
\end{algorithmic}
\caption{Algorithmic description of COMIX. 
	The function $\text{CEM}$ is defined in Algorithm \ref{alg:cem}.} 
\label{alg:comix}
\end{algorithm}

\section{Environment Details}
\label{sec:env_details}
\subsection{Continuous Predator-Prey}
\label{sec:pp_sup}
\begin{figure}[ht!]
   \begin{center}
    \includegraphics[width=0.6\textwidth]{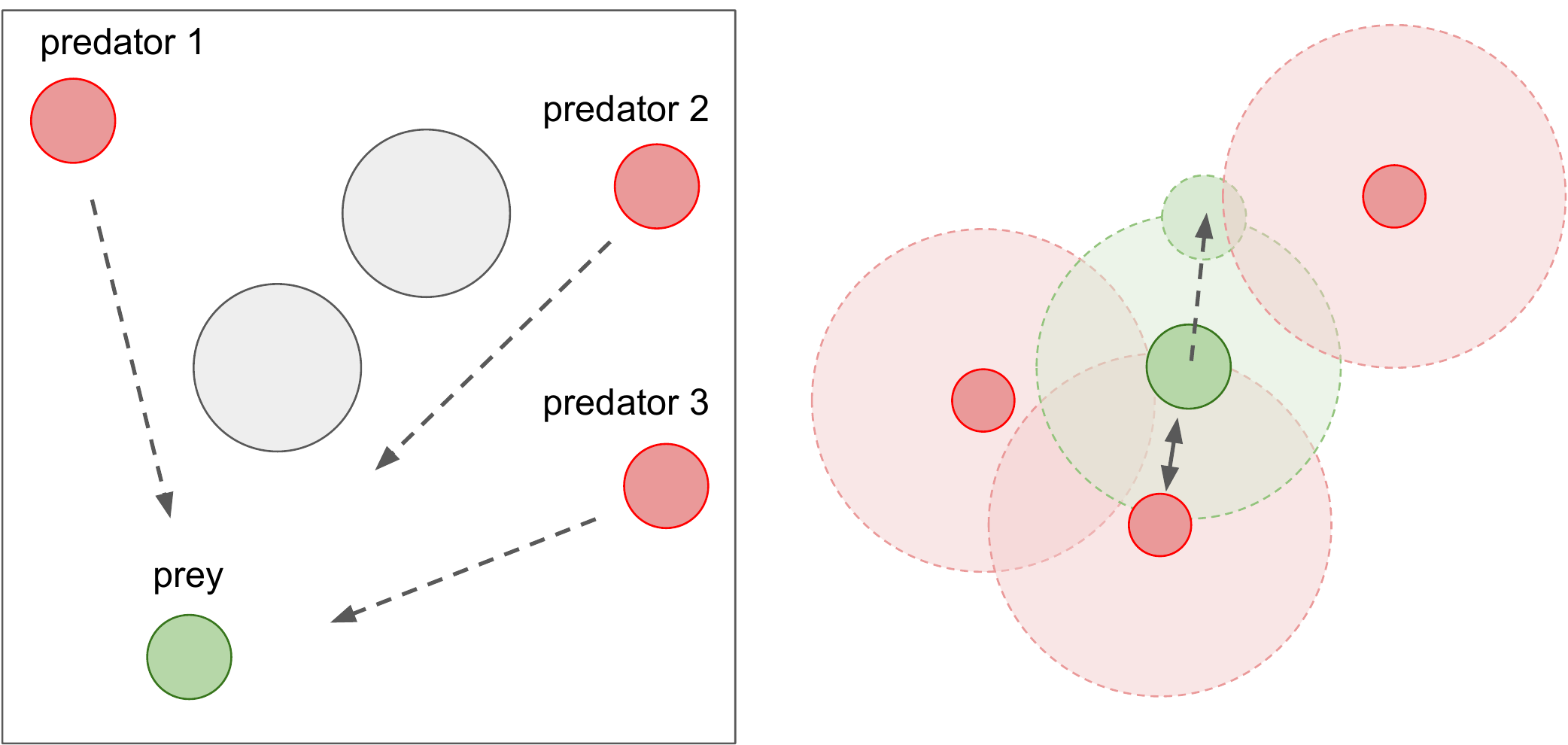}
    \end{center}
    \caption{Continuous Predator-Prey. \textbf{Left: }Top-down view of toroidal plane, with predators (red), prey (green) and obstacles (grey). \textbf{Right: }Illustration of the prey's avoidance heuristic. Observation radii of both agents and prey are indicated. }
    \label{fig:multiagent_particle}
\end{figure}

We consider the mixed \textit{simple tag} environment (Figure \ref{fig:multiagent_particle}) introduced by \citet{lowe2017multi}, which is a variant of the classic predator-prey game. Three slower cooperating circular agents (red), each with continuous movement action spaces $u^a\in\mathbb{R}^2$, must catch a faster circular prey (green) on a randomly generated two-dimensional toroidal plane with two large landmarks blocking the way. 

To obtain a purely cooperative environment, we replace the prey's policy by a hard-coded heuristic, that, at any time step, moves the prey to the sampled position with the largest distance to the closest predator. If one of the cooperative agents collides with the prey, a team reward of $+10$ is emitted; otherwise, no reward is given. In the original simple tag environment, each agent can observe the relative positions of the other two agents, the relative position and velocity of the prey, and the relative positions of the landmarks. This means each agent's private observation provides an almost complete representation of the true state of the environment. 

To introduce partial observability to the environment, we add an agent \textit{view radius}, which restricts the agents from receiving information about other entities (including all landmarks, the other two agents, and the prey) that are out of range. Specifically, we set the view radius such that the agents can only observe other agents roughly $60\%$ of the time. We open-source the full set of multi-agent particle environments with added partial observability.
\footnote{\url{https://github.com/schroederdewitt/multiagent-particle-envs/.}}

In addition, we implement a variant of our Continuous Predator-Prey task (with 3 agents and 1 prey), where the reward function is modified to make the task \textit{nonmonotonic}. Specifically, if one agent collides with the prey while at least another one being close enough, a team reward of $+10$ is given. However, if only one agent collides with the prey without any other agent being close enough, a negative team reward of $-1$ is given. Otherwise, no reward is provided.

\subsection{Multi-Agent MuJoCo} 
All MAMuJoCo environments we tested are configured according to its default configuration, where each agent can observe only positions (not velocities) of its own body parts and at graph distances greater than zero. 
In ManyAgent Swimmer, we configure the number of agents to be $10$, each controlling a consecutive segment of length $2$. We thus refer to this environment as ManyAgent Swimmer [10x2]. We set maximum observation distances to $k=0$ for all three environments tested, including 2-Agent Humanoid, 2-Agent HumanoidStandup, and ManyAgent Swimer [10x2]. Default team reward is used (see Table \ref{tab:multiagent_mujoco_goals}).

\subsection{SMAC} 
SMAC consists of a set of complex StarCraft II micromanagement tasks that are carefully designed to study decentralised multi-agent control.
The tasks in SMAC involve combat between two armies of units. The first army is controlled by a group of learned allied agents. The second army consists of enemy units controlled by the built-in heuristic  AI. The goal of the allied agents is to defeat the enemy units in battle, to maximise the win rate. The action space consists of a set of discrete actions: $\mathtt{move}$ in four cardinal directions, $\mathtt{attack}$ any selected enemy (available if the enemy is within the agent's shooting range), $\mathtt{stop}$, and $\mathtt{noop}$. Hence the number of actions increases as the number of enemies increases. All experiments on SMAC use the default reward and observation settings of the SMAC benchmark \cite{samvelyan19smac}.

\section{Experimental Details}
\label{sec:exp_details}
We evaluate the performance of each method
using the following procedure: for each run of a method, we pause training every fixed number of timesteps ($2000$ timesteps for Continuous Predator-Prey, $4000$ timesteps for MAMuJoCo, and $10000$ timesteps for SMAC) and run a fixed number of independent episodes ($10$ episodes for Continuous Predator-Prey and MAMuJoCo, and $32$ episodes for SMAC) with each agent performing action selection greedily in a decentralised fashion. On both Continuous Predator-Prey and MAMuJoCo, the mean value of these episode returns are used to evaluate the performance of the learned policies. On SMAC, we use the median test win rate (i.e., the percentage of the $32$ episodes where the agents defeat all enemy units within the permitted time limit) to evaluate the learned policies, as in \cite{samvelyan19smac}. 

\subsection{Continuous Predator-Prey}
In value-based methods COVDN and COMIX, the architecture of the shared agent network is a DRQN with a recurrent layer comprised of a GRU with a $64$-dimensional hidden state, with a fully-connected layer before and after.
In actor-critic methods \name{}, \name{}-vdn, MADDPG, and IDDPG, the architecture of the shared agent network is also a DRQN with a recurrent layer comprised of a GRU with a $64$-dimensional hidden state, with a fully-connected layer before and after, while the final output layer is a tanh layer, to bound actions. The shared critic network is a MLP with $2$ hidden layers of $64$ units and ReLU non-linearities. 
All agent networks receive the current local observation and last individual action as input. 
In MADDPG, the centralised critic receives the global state and the joint action of all agents as input. The global state consists of the joint observations of all agents in Continuous Predator-Prey. In other actor-critic methods, there is a shared critic network that approximates per-agent utilities, which receives each agent's local observation and individual action as input.

During training and testing, we restrict each episode to have a length of $25$ time steps. Training lasts for $2$ million timesteps. 
To encourage exploration, we use uncorrelated, mean-zero Gaussian noise with $\sigma=0.1$ during training (for all $2$ million timesteps).
We set $\gamma=0.85$ for all experiments. The replay buffer contains the most recent $10^6$ transitions. We train on a batch size of $1024$ after every timestep. For the soft target network updates we use $\tau = 0.001$. All neural networks (actor and critic) are trained using Adam optimiser with a learning rate of $0.01$. To evaluate the learning performance, the training is paused after every $2000$ timesteps during which $10$ independent test episodes are run with agents performing action selection greedily in a decentralised fashion. 

\subsection{Multi-Agent MuJoCo}
In all value-based methods, the architecture of all agent networks is a MLP with $2$ hidden layers with $400$ and $300$ units respectively, similar to the setting used in OpenAI Spinning Up.\footnote{\url{https://spinningup.openai.com/en/latest/}.} All agent networks use ReLU non-linearities for all hidden layers. In all actor-critic methods, the architecture of the shared agent network and critic network is also a MLP with $2$ hidden layers with $400$ and $300$ units respectively, while the final output layer of the actor network is a tanh layer, to bound the actions. In all value-based methods, the agent receives its current local observation as input. 
In MADDPG, the centralised critic receives the global state and the joint action of all agents as input. The global state consists of the full state information returned by the original OpenAI Gym \cite{brockman_openai_2016}. 
In other actor-critic methods, there is a shared critic network that approximates per-agent utilities, which receives each agent's local observation and individual action as input.

During training and testing, we restrict each episode to have a length of $1000$ time steps. Training lasts for $2$ million or $4$ million timesteps. 
To encourage exploration, we use uncorrelated, mean-zero Gaussian noise with $\sigma=0.1$ during training. 
We also use the same trick as in OpenAI Spinning Up to improve exploration at the start of training. For a fixed number of steps at the beginning (we set it to be $10000$), the agent takes actions which are sampled from a uniform random distribution over valid actions. After that, it returns to normal Gaussian exploration. We set $\gamma=0.99$ for all experiments. The replay buffer contains the most recent $10^6$ transitions. We train on a batch size of 100 after every timestep. For the soft target network updates we use $\tau = 0.001$. All neural networks (actor and critic) are trained using Adam optimiser with a learning rate of $0.001$. To evaluate the learning performance, the training is paused after every $4000$ timesteps during which $10$ independent test episodes are run with agents performing action selection greedily in a decentralised fashion.

\subsection{SMAC}
For baseline algorithms DOP \cite{wang2021dop}, COMA \cite{foerster_counterfactual_2018}, CentralV \cite{foerster_counterfactual_2018}, QMIX \cite{rashid2020monotonic}, and QPLEX \cite{wang2020qplex}, we use the the same training setup as provided by their authors where the hyperparameters have been fine-tuned on the SMAC benchmark.

Most of our training hyperparameters for \name{} and MADDPG \cite{lowe2017multi} follow \cite{rashid2020monotonic}. 
In both methods, the architecture of the shared actor network is a DRQN with a recurrent layer comprised of a GRU with a $64$-dimensional hidden state, with a fully-connected layer before and after. The shared critic network is a MLP with $2$ hidden layers of $64$ units and ReLU non-linearities.
Exploration is performed during training using a scheme similar to COMA \cite{foerster_counterfactual_2018}.
Action probabilities are produced from the final layer of the actor network, $z$, via a bounded softmax distribution that lower-bounds the probability of any given action by $\epsilon/|U|$: $P(u) = (1 - \epsilon)\mathrm{softmax}_u + \epsilon/|U|$, where $|U|$ is the size of the joint action space. Throughout the training, we anneal $\epsilon$ linearly from $0.5$ to $0.05$ over $50k$ timesteps and keep it constant for the rest of the training. The replay buffer contains the most recent $5000$ episodes.
We sample batches of $32$ episodes uniformly from the replay buffer and train on fully unrolled episodes.
In MADDPG, we use a target network for the actor and critic, respectively.
In \name{}, we use a target network for the actor, critic, and mixing network, respectively.
All target networks are periodically updated every $200$ training steps. 
All neural networks are trained using Adam optimiser with learning rate $0.0025$ for the actor network and $0.0005$ for the critic network. We set $\gamma=0.99$ for all experiments. 

The architecture of the mixing network in \name{} follows \cite{rashid2020monotonic}. It consists of a single hidden layer of $32$ units with an ELU non-linearity. 
The weights of the mixing network are produced by separate hypernetworks.
The hypernetworks consist of a feedforward network with a single hidden layer of $64$ units with a ReLU non-linearity. 
The output of the hypernetwork is passed through an absolute activation function (to acheive non-negativity) and then resized into a matrix of appropriate size.

\section{Additional Results on Different Critic Factorisations}
\label{sec:nonmon_fac}

\begin{figure}[ht!]
    \centering
    \includegraphics[width=0.45\columnwidth]{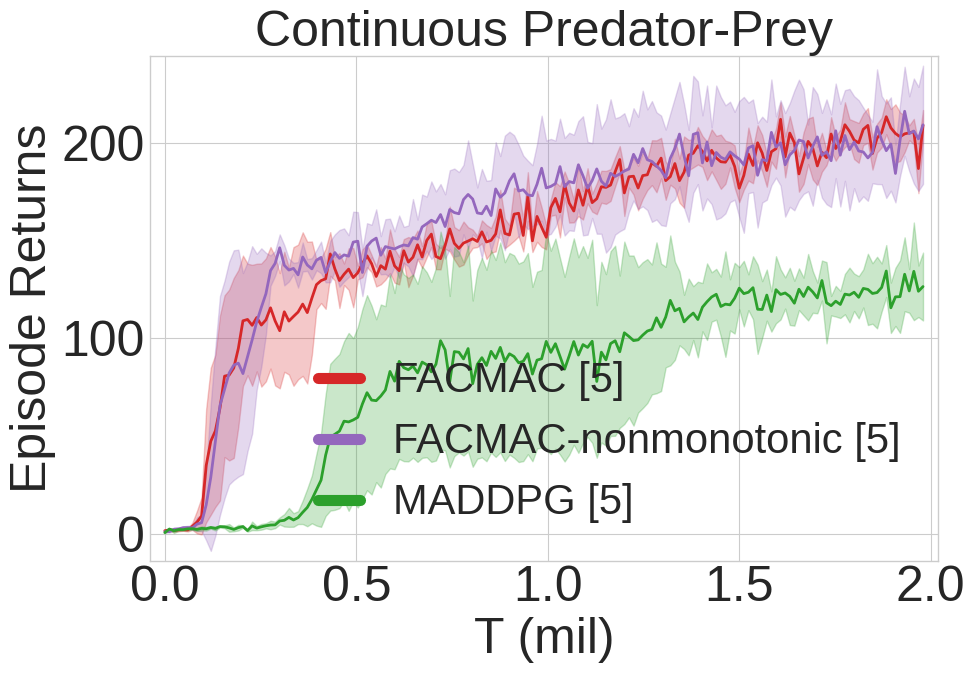}
    \caption{Mean episode return on our Continuous Predator-Prey task (with $3$ agents and $1$ prey).}
    \label{fig:cts_pp_nonmon}
\end{figure}

\begin{figure*}[ht!]
	\centering
	\subfigure[{Easy}]{
		\label{fig:smac_2s3z_pg}
		\includegraphics[width=0.3\textwidth]{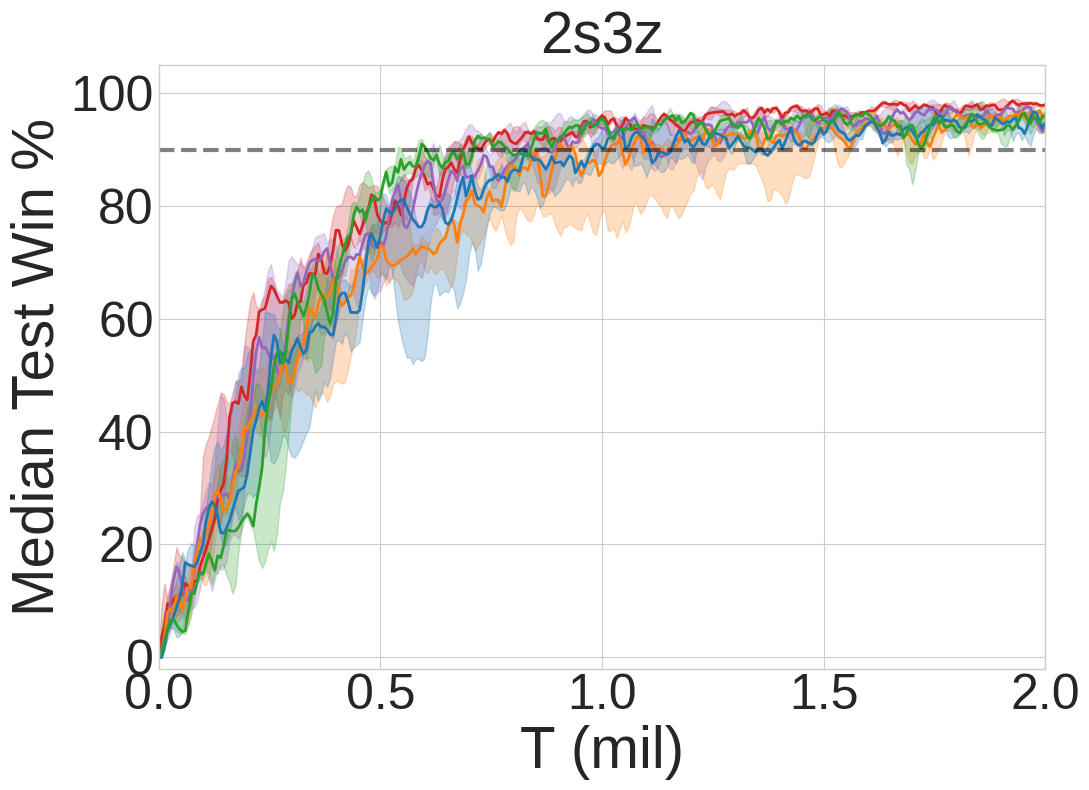}
	}
	\subfigure[{Easy}]{
		\label{fig:smac_mmm_pg}
		\includegraphics[width=0.3\textwidth]{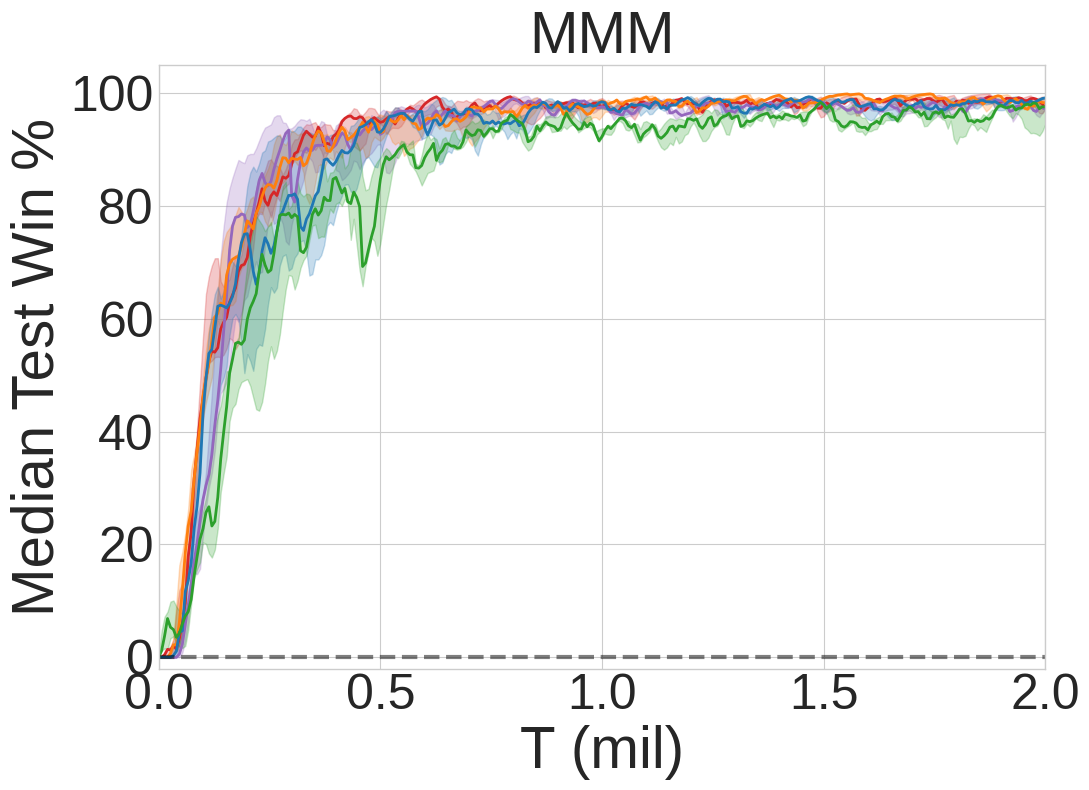}
	}
	\subfigure[{Hard}]{
		\label{fig:smac_2c_vs_64zg_pg}
		\includegraphics[width=0.3\textwidth]{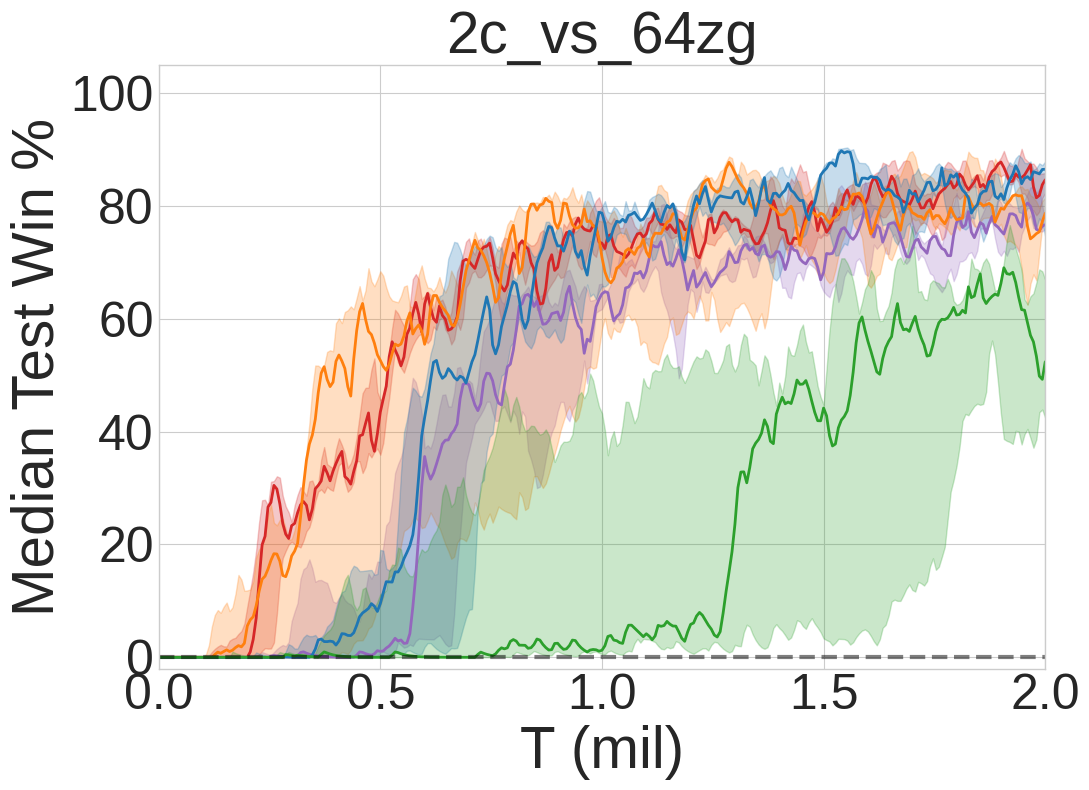}
	}
	\subfigure[{Hard}]{
		\label{fig:smac_bane_vs_bane_pg}
		\includegraphics[width=0.3\textwidth]{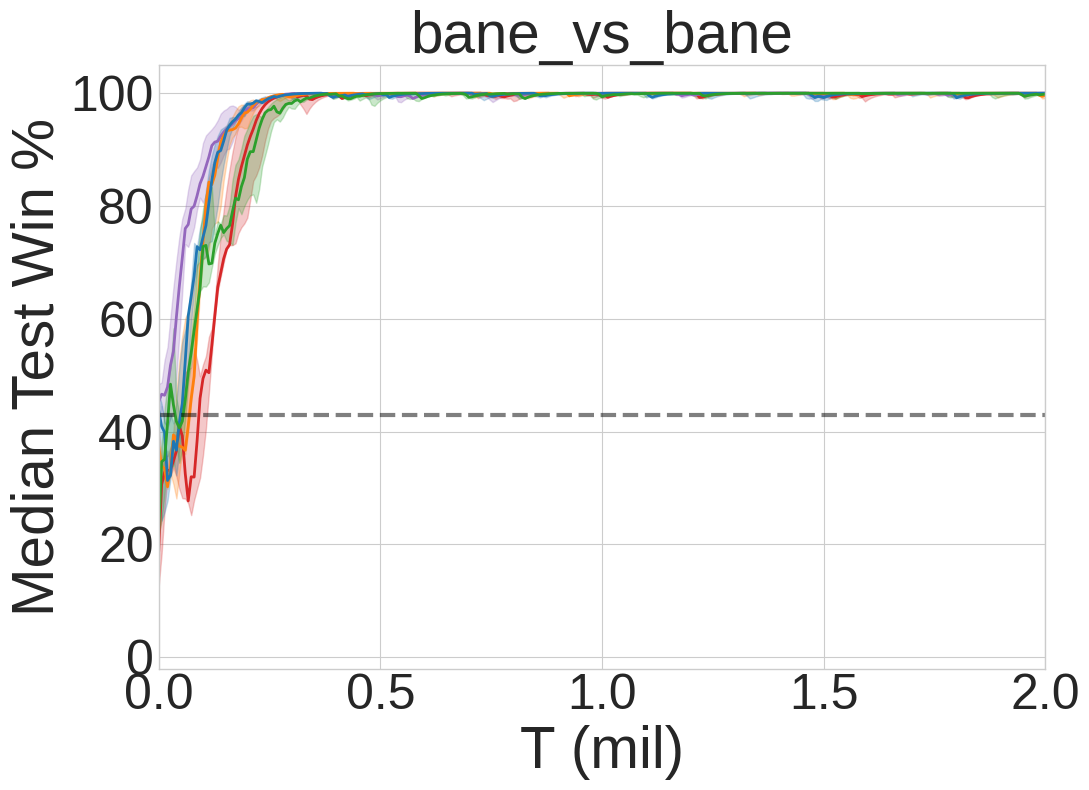}
	}
	\
	\subfigure[{Super Hard}]{
		\label{fig:smac_mmm2_pg}
		\includegraphics[width=0.3\textwidth]{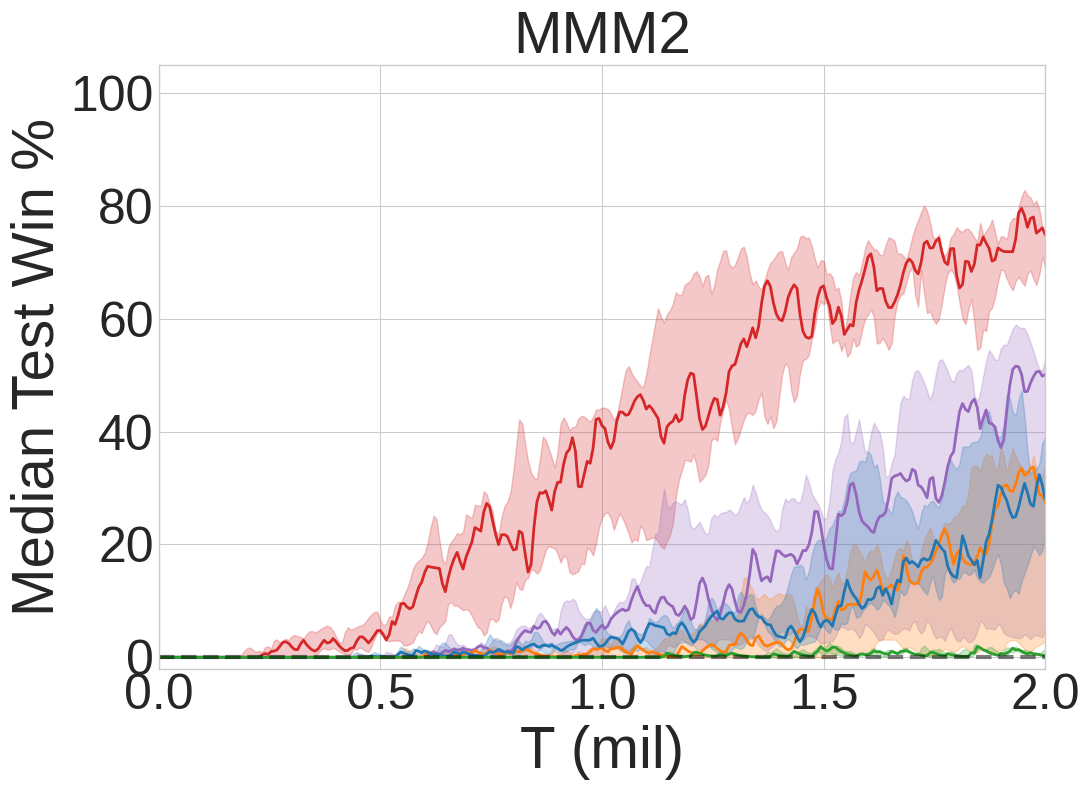}
	}
	\subfigure[{Super Hard}]{
		\label{fig:smac_27m_vs_30m_pg}
		\includegraphics[width=0.3\textwidth]{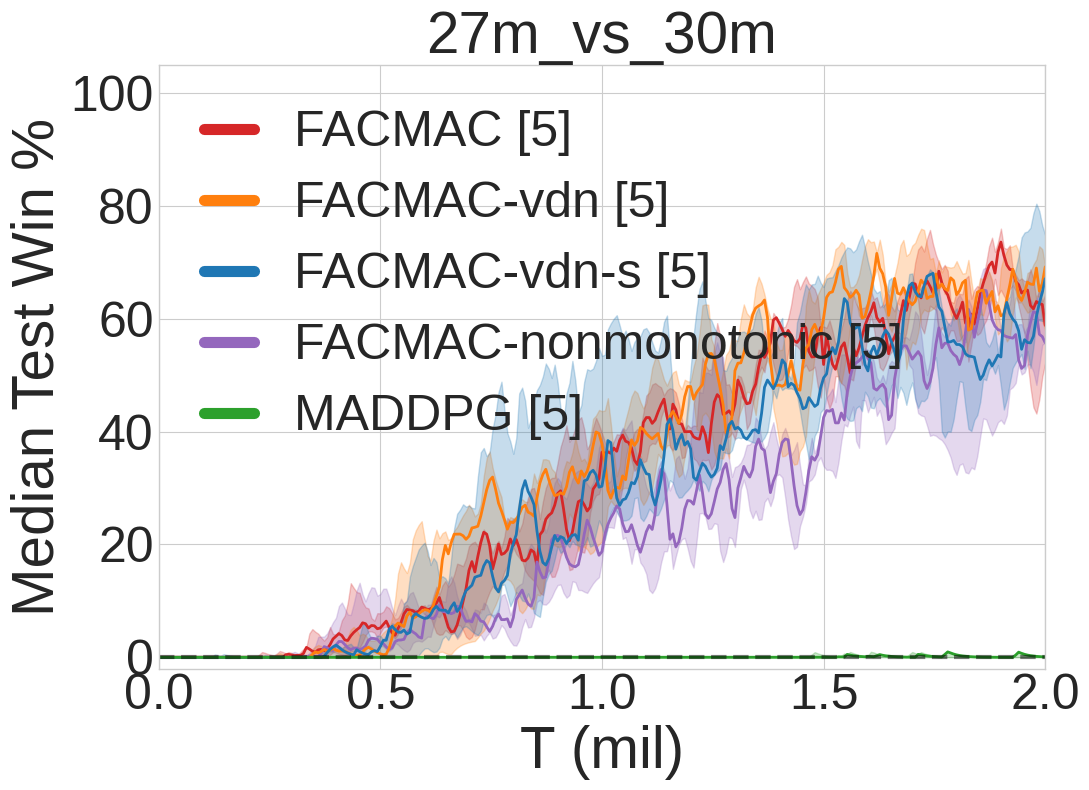}
	}
	\caption{Median test win \% on six different SMAC maps: (a) \textit{2s3z} (easy), (b) \textit{MMM} (easy), (c) \textit{2c\_vs\_64zg} (hard), (d) \textit{bane\_vs\_bane} (hard), (e) \textit{MMM2} (super hard), and (f) \textit{27m\_vs\_30m} (super hard), comparing \name{} with different forms of critic factorisations.}
	\label{fig: smac_results_diff_factorisations}
\end{figure*}

\end{document}